\documentclass{article} %
\usepackage{iclr2023_conference,times}
\PassOptionsToPackage{numbers, compress}{natbib}
\usepackage{natbib}

\usepackage[utf8]{inputenc} %
\usepackage{url}
\usepackage{color}
\usepackage{booktabs}       %
\usepackage{amsfonts}       %
\usepackage{nicefrac}       %
\usepackage{microtype}      %

\usepackage{microtype}
\usepackage{graphicx}
\usepackage{booktabs} %
\usepackage{comment}
\usepackage{amsmath}

\usepackage{amsmath}
\usepackage{amssymb}
\usepackage{mathtools}
\usepackage{amsthm}
\usepackage{comment}
\usepackage{array}
\usepackage{float}
\floatstyle{plaintop}
\restylefloat{table}

\usepackage[algo2e,linesnumbered,noend]{algorithm2e}
\usepackage{mathtools}
\usepackage{adjustbox}

\usepackage{algorithm,algpseudocode,setspace}
\usepackage{eqnarray}

\usepackage{wrapfig}
\usepackage{transparent}

\usepackage[font=small,labelfont=bf]{caption}
\usepackage{subcaption}

\newcommand{\nRuns}{\debug T}

\newtheorem{theorem}{Theorem}		%
\newtheorem{corollary}{Corollary}		%
\newtheorem{lemma}{Lemma}		%
\newtheorem{definition}{Definition}
\newtheorem{proposition}{Proposition}		%

\newtheorem*{corollary*}{Corollary}		%

\DeclarePairedDelimiter{\norm}{\lVert}{\rVert}

\usepackage[T1]{fontenc}    %
\usepackage{xr-hyper}
\usepackage[hypertexnames=false]{hyperref}      
\usepackage{xcolor}
\hypersetup{
    colorlinks,
    linkcolor={red!50!black},
    citecolor={blue!50!black},
    urlcolor={blue!80!black}
}

\numberwithin{equation}{section}		%
\usepackage[sort&compress,capitalize,nameinlink]{cleveref}		%
\usepackage[
    type={CC},
    modifier={by},
    version={3.0},
]{doclicense}
\crefname{assumption}{Assumption}{Assumptions}

\usepackage[textwidth=30mm]{todonotes}		%

\newcommand{\debug}[1]{#1}		%

\theoremstyle{definition}
\newtheorem{assumption}{Assumption}		%
\newtheorem{example}{Example}		%

\newtheorem*{definition*}{Definition}		%
\newtheorem*{assumption*}{Assumptions}		%

\theoremstyle{remark}
\newtheorem{remark}{Remark}		%

\newtheorem*{remark*}{Remark}		%
\newtheorem*{example*}{Example}		%

\newcounter{proofpart}

\numberwithin{example}{section}		%

\newcommand\blfootnote[1]{%
  \begingroup
  \renewcommand\thefootnote{}\footnote{#1}%
  \addtocounter{footnote}{-1}%
  \endgroup
}

\DeclarePairedDelimiterX{\setdef}[2]{\{}{\}}{#1:#2}		%
\DeclarePairedDelimiterXPP{\exclude}[1]{\mathopen{}\setminus}{\{}{\}}{}{#1}

\newcommand{\eg}{e.g.,\xspace}		%
\newcommand{\ie}{i.e.,\xspace}		%

\DeclareMathOperator{\bigoh}{\mathcal O}		%
\DeclareMathOperator{\sign}{sgn}		%
\newcommand{\vdim}{\debug d}		%

\DeclarePairedDelimiterX{\braket}[2]{\langle}{\rangle}{#1,#2}		%
\DeclareMathOperator{\gap}{Gap}         %

\newcommand{\obj}{\debug f}		%
\newcommand{\smooth}{\debug \beta}		%

\DeclareMathOperator{\val}{val}		%

\DeclareMathOperator{\ex}{\mathbb{E}}		%
\DeclareMathOperator{\prob}{\mathbb{P}}		%

\DeclarePairedDelimiterXPP{\exof}[1]{\ex}{[}{]}{}{%
 #1}

\DeclarePairedDelimiterXPP{\probof}[1]{\prob}{(}{)}{}{%
 #1}

\DeclareMathOperator{\one}{\mathbbm{1}}		%

\DeclarePairedDelimiterX{\product}[2]{\langle}{\rangle}{#1,#2}		%
\DeclarePairedDelimiterXPP{\dnorm}[1]{}{\lVert}{\rVert}{_{\ast}}{#1}		%

\newcommand{\ud}{\,\mathrm{d}}
\newcommand{\ENCODE}{\mathrm{CODE\circ Q}}
\newcommand{\DECODE}{\mathrm{DEQ\circ CODE}}

\newcolumntype{C}[1]{>{\centering\arraybackslash}m{#1}}

\def\reals{\mathbb{R}}

\def\integers{\mathbb{Z}}

\def\ebf{{\bf e}}

\def\qbf{{\bf q}}

\def\sbf{{\bf s}}

\def\ubf{{\bf u}}
\def\vbf{{\bf v}}

\def\xbf{{\bf x}}

\def\zbf{{\bf z}}

\def\xbf{{\bf x}}

\def\Bc{{\cal B}}
\def\Cc{{\cal C}}

\def\Fc{{\cal F}}

\def\Lc{{\cal L}}

\def\Oc{{\cal O}}

\def\Sc{{\cal S}}

\def\Uc{{\cal U}}

\def\Xc{{\cal X}}

\def\Zc{{\cal Z}}

\def\nn{\nonumber}
\def\beq{\begin{equation}}
\def\eeq{\end{equation}}
\def\beqa{\begin{eqnarray}}
\def\eeqa{\end{eqnarray}}
\def\balign{\begin{align}}
\def\ealign{\end{align}}
\def\bpr{\begin{proof}}
\def\epr{\end{proof}}
\def\bth{\begin{theorem}}
\def\eth{\end{theorem}}
\def\blm{\begin{lemma}}
\def\elm{\end{lemma}}
\def\bprop{\begin{proposition}}
\def\eprop{\end{proposition}}
\def\bcr{\begin{corollary}}
\def\ecr{\end{corollary}}
\def\ie{{\it i.e.,\ \/}}
\def\eg{{\it e.g.,\ \/}}

\def\E{\mathbb{E}}

\def\and {{\rm and}}

\newcommand{\cdf}{F}
\newcommand{\ql}{\ell}
\newcommand{\bql}{{\boldsymbol \ell}}
\newcommand{\level}[1]{{\tau(#1)}}

\newcommand{\qcoeff}{{\xi}}
\newcommand{\feasl}{{\Lc}}

\newcommand{\signvec}{{\sbf}}

\newcommand{\EG}{{EG}\xspace}

\newcommand{\QEG}{{Q-GenX}\xspace}

\newcommand{\sgd}{{SGD}\xspace}

\newcommand{\qinf}{{QSGDinf}\xspace}

\newcommand{\nuq}{{NUQSGD}\xspace}

\newcommand{\mathbbm}[1]{\text{\usefont{U}{bbm}{m}{n}#1}} %

\def \l{\left}
\def \r{\right}
\def \a{\alpha}

\def \g{\gamma}

\def\o{\omega}

\def \s{\sigma}

\def \ol{\overline}

\def\ra{\rightarrow}

\title{Distributed Extra-gradient with Optimal\\ Complexity and Communication Guarantees}

\author{Ali Ramezani-Kebrya\thanks{These authors contributed equally to this work.}~~\thanks{Department of Informatics, University of Oslo. Work performed at EPFL (LIONS) and Aalborg University.}~~ Kimon Antonakopoulos\footnotemark[1]~~\thanks{Laboratory for Information and Inference Systems (LIONS), EPFL.}~~ Igor Krawczuk\footnotemark[1]~~\footnotemark[3]~~ Justin Deschenaux\footnotemark[1]~~\footnotemark[3]\\ {\hspace{5.5cm}\bf Volkan Cevher\footnotemark[3]}\\
\hspace{2cm} $^{\dagger}$\href{mailto:ali@uio.no}{\texttt{ali@uio.no}}\hspace{1cm}  $^{\ddagger}$\href{mailto:firstname.lastname@epfl.ch}{\texttt{firstname.lastname@epfl.ch}}
}

\iclrfinalcopy %
\begin{document}

\maketitle
\begin{abstract}
\blfootnote{Open source code  will be released at \url{https://github.com/LIONS-EPFL/qgenx}}
We consider monotone variational inequality (VI) problems in {\it multi-GPU}  settings where multiple processors/workers/clients have access to 
local {\it stochastic dual vectors}. This setting  includes a broad range of important problems from distributed convex minimization to min-max and games. Extra-gradient, which is a de facto algorithm  for monotone VI problems, has not been designed to be communication-efficient. To this end, we propose a quantized generalized extra-gradient (\QEG), which is an unbiased and adaptive compression method tailored to solve VIs.
We provide an adaptive step-size rule, which  adapts to the respective noise profiles at hand and achieve a fast rate of  $\Oc(1/T)$ under relative noise, and an order-optimal $\Oc(1/\sqrt{T})$ under absolute noise  and show distributed training accelerates convergence. Finally, we validate our theoretical results by providing real-world experiments and training generative adversarial networks on multiple GPUs.

\end{abstract}
\section{Introduction}
\label{sec:intro}
The surge of deep learning across tasks beyond image classification has triggered a vast literature of optimization paradigms, which transcend the standard empirical risk minimization. For example, training generative adversarial networks (GANs) gives rise to solving a more complicated zero-sum game between a generator and a discriminator~\citep{goodfellow2020generative}. This can become even more complex when the generator and the discriminator do not have completely antithetical objectives and e.g., constitute a more general game-theoretic setup.
A powerful unifying framework which includes those important problems as special cases is {\it monotone variational inequality (VI)}. 
Formally, 
given a monotone operator $A:\reals^{d}\to \reals^{d}$, \ie
\begin{equation}\nn
    \braket{A(\xbf)-A(\xbf')}{\xbf-\xbf'}\geq 0\;\;\text{for all}\;\;\xbf,\xbf'\in \reals^{d},
\end{equation}
our goal is to find some $\xbf^*\in\reals^d$ such that:
\begin{align}
\tag{VI}
\label{eq:VI}
    \langle A(\xbf^*),\xbf-\xbf^* \rangle \geq 0,\;\;\text{for all} \;\;\xbf\in\reals^d.
\end{align}

Several practical problems can be formulated as a \eqref{eq:VI} problem including  those with convex-like structures, \eg   
{\it convex minimization}, {\it saddle-point problems}, and {\it games}~\citep{facchinei2003finite,BC17,Universal}  with several applications such as auction theory~\citep{syrgkanis2015fast}, multi-agent and robust reinforcement learning~\citep{pinto2017robust}, adversarially robust
learning~\citep{schmidt2018adversarially}, and GANs.

For various tasks, it is widely known that employing {\it deep neural networks} (DNNs) along with massive datasets leads to significant improvement in terms of learning~\citep{shalev2014understanding}. However, DNNs can no longer be trained on a single machine. One common solution is to train on multi-GPU systems~\citep{QSGD}. Furthermore, in federated learning (FL),  multiple clients, \eg a few hospitals  or several cellphones learn a model collaboratively without sharing local data due to privacy risks~\citep{FL}.

To minimize a single empirical risk, \sgd is the most popular algorithm due to its flexibility for parallel implementations and excellent generalization performance~\citep{QSGD,wilson2017marginal}. Data-parallel \sgd has delivered tremendous success in terms of scalability: \citep{Zinkevich,Scaleup,Recht11,Dean12,Coates,Projectadam,Li14,Duchi15,Petuum,Zhang15,QSGD,ALQ,NUQSGD,FL}. Data-parallel \sgd reduces computational costs significantly. However, the communication costs for broadcasting huge stochastic gradients is the main performance bottleneck in large-scale settings \citep{Strom15,QSGD,ALQ,NUQSGD,FL}.  %

Several methods have been proposed to accelerate training for classical empirical risk minimization  such as gradient (or model update) compression, gradient sparsification, weight quantization/sparsification, and reducing the frequency of communication though local methods~\citep{Dean12,Seide14,Sa,Gupta,Abadi,QSGD,TernGrad,Dorefa,signSGD,ALQ,NUQSGD,FL}. In particular, {\it unbiased gradient quantization} is interesting due to both enjoying strong theoretical guarantees along with providing communication efficiency  on the fly, \ie convergence under the same hyperparameteres tuned for uncompressed variants while providing substantial savings in terms of communication costs~\citep{QSGD,ALQ,NUQSGD}. 
Unlike full-precision data-parallel \sgd, where each processor is required to broadcast its local gradient in full-precision, 
\ie transmit and receive huge full-precision vectors at each iteration, unbiased quantization requires each processor to transmit only a few communication bits per iteration for each component of the stochastic gradient.

In this work, we propose communication-efficient variants of a general first-order method that achieves the {\it optimal rate of convergence} with {\it improved guarantees on the number of communication bits} for monotone~\ref{eq:VI}s and {\it show distributed training accelerates convergence}. We employ an adaptive step-size and both adaptive and non-adaptive variants of unbiased quantization schemes tailored to~\ref{eq:VI}s. 

There exist three major challenges to tackle this
problem: 1) how to modify adaptive variants of unbiased quantization schemes tailored to solve general~\ref{eq:VI}s; 2) can we achieve optimal rate of convergence without knowing noise profile and show benefits of distributed training?; 3) can we validate improvements in terms of scalability without compromising accuracy in large-scale settings? We aim to address those challenges and answer all questions in the affirmative: 

\subsection{Summary of contributions} 

\begin{itemize}
\item We  propose  {\it quantized} generalized extra-gradient (\QEG) family of algorithms, which employs unbiased compression methods tailored to general \ref{eq:VI}-solvers. Our framework unifies distributed and communication-efficient variants of stochastic dual averaging, stochastic dual extrapolation, and stochastic optimistic dual averaging.

\item Without prior knowledge on the noise profile, we provide an adaptive step-size rule for \QEG and achieve a fast rate of  $\Oc(1/T)$ under relative noise, and an order-optimal $\Oc(1/\sqrt{T})$ in the absolute noise case and show that increasing the number of processors accelerates convergence for general monotone \ref{eq:VI}s.

\item We validate our theoretical results by providing real-world experiments and training generative adversarial networks on multiple GPUs.%

\end{itemize}

\subsection{Related work}\label{sec:related}
We overview a summary of related work. Complete related work is provided in~\cref{app:relatedwork}.
Adaptive quantization has been used for speech communication  \citep{Cummiskey}. In machine learning, adapting quantization levels~\citep{ALQ}, 
adapting communication frequency in local SGD~\citep{wang2019adaptive}, adapting the number of quantization levels (communication budget) over the course of training~\citep{guo2020accelerating,agarwal2021adaptive},~adapting a gradient sparsification scheme over the course of training~\citep{khirirat2021flexible}, and adapting compression parameters across
model layers and training iterations~\citep{CGX} have been proposed for {\it minimizing a single empirical risk}. In this paper, we propose  communication-efficient generalized extra-gradient family of algorithms with  adaptive quantization and adaptive step-size for a general \eqref{eq:VI} problem. In the \ref{eq:VI}~literature, the benchmark method is  \EG, proposed by \citet{Kor76}, along with its variants including~ \citep{Nem04,Nes07}. Rate interpolation between different noise profiles under an adaptive step-size has been explored by \citet{Universal}. However, their results are limited to {\it centralized and single-GPU settings}.~\citet{beznosikov2021distributed,kovalev2022optimal} have proposed communication-efficient algorithms for \ref{eq:VI}s with finite-sum structure and variance reduction in centralized settings and (strongly) monotone \ref{eq:VI}s in decentralized settings, respectively. Unlike \citep{beznosikov2021distributed,kovalev2022optimal}, we achieve fast and order-optimal rates with {\it adaptive step-size} and {\it adaptive compression}  without requiring variance reduction and strong monotoncity, and improve variance and code-length bounds for unbiased and adaptive compression.

\section{Problem setup}
\label{sec:prelims}
Our objective throughout this paper is to solve \ref{eq:VI} with $A:\reals^{d}\to \reals^{d}$ being a monotone operator\footnote{Notations are provided in~\cref{app:init}.}.

Moreover, in order to avoid trivialities, we make the following mild assumption:
\begin{assumption}[Existence]\label{assumption:ex}
The set $\Xc^*:=\{\xbf^*\in\reals^d:\xbf^*~\text{solves}~\eqref{eq:VI}\}$ is non-empty. 
\end{assumption}
Let $\Cc$ denote a non-empty compact test domain. A popular performance measure for the evaluation of a candidate solution for \ref{eq:VI} is the so-called \emph{restricted gap function} defined as:
\begin{equation}
\label{eq:gap}
\tag{Gap}
\gap_{\Cc}(\hat\xbf)=\sup_{\xbf\in \Cc}\langle A(\xbf),\hat\xbf-\xbf\rangle.
\end{equation}
\ref{eq:gap} is used to measure $\hat\xbf$'s performance  mainly because it characterizes the solutions of \ref{eq:VI} via its zeros. Mathematically speaking, we have the following proposition:
\begin{proposition}[\citealt{Nes09}]
\label{proposition:gap}
Let $\Cc$ be a non-empty convex subset of $\reals^{d}$. Then, the following holds 
\begin{enumerate}
\item
$\gap_{\Cc}(\hat{\xbf})\geq 0$~for all $\hat{\xbf}\in \Cc$
\item
If $\gap_{\Cc}(\hat{\xbf})=0$ and $\Cc$ contains a neighbourhood of $\hat{\xbf}$, then $\hat{\xbf}$ is a solution of \ref{eq:VI}. 
\end{enumerate} 
\end{proposition} 
\cref{proposition:gap} is an extension of  an earlier characterization shown by \citet{Nes07}; we refer the reader to \citep{ABM19,Nes09} and references therein.

From an algorithmic perspective, we primarily consider the generic family of iterative methods, which have access to a {\it stochastic first-order} oracle, \ie a black-box feedback mechanism \citep{Nes04}. The respective iterative algorithm can call the oracle over and over at a (possibly) random sequence of points $\xbf_0,\xbf_1,\ldots$ When called at $\xbf$, the oracle draws an i.i.d. sample $\o$ from a (complete) probability space $(\Omega,\Fc,\mathbb{P})$ and returns a  {\it stochastic dual vector} $g(\xbf;\o)$ given by 
\begin{align}
\label{eq:oracle}
 g(\xbf;\o)=A(\xbf)+U(\xbf;\o)
\end{align} 
where $U(\xbf;\o)$ denotes the measurement error or noise. We consider two important noise profile models, i.e.,  {\it absolute} and {\it relative} noise models formally described in the following assumptions:

\begin{assumption}
[Absolute noise]
\label{assumption:random}
Let $\xbf\in\reals^d$ and $\o\sim\mathbb{P}$. The oracle $g(\xbf;\o)$ enjoys these properties: 1)
\emph{Almost sure boundedness:} There exists some  $M>0$ such that  $\|g(\xbf;\o)\|_{\ast}\leq M$ a.s.; 2) \emph{Unbiasedness:} $\E\l[g(\xbf;\o)\r]=A(\xbf)$; 3) \emph{Bounded absolute variance:}
$\E\l[\norm{U(\xbf;\o)}_{\ast}^{2}\r]\leq \sigma^{2}$.
\end{assumption}
The conditions in \cref{assumption:random} are mild and hold for standard oracles, in particular, in the context of adaptive algorithms~\citep{KLBC19,LYC18,BL19,AM21} and typically guarantee a convergence rates in $\bigoh(1/\sqrt{T})$ \citep{NJLS09,JNT11,Universal}.
Alternatively,  one may consider the \emph{relative} noise model following \citep{Pol87}:
\begin{assumption}[Relative noise]
\label{assumption:relative}
The oracle $g(\xbf;\o)$ satisfies: 
1) \emph{Almost sure boundedness:} There exists some  $M>0$ such that  $\|g(\xbf;\o)\|_{\ast}\leq M$ a.s; 2)
\emph{Unbiasedness:} $\E\l[g(\xbf;\o)\r]=A(\xbf)$; 3) 
\emph{Bounded relative variance:}
There exists some  $c>0$ such that
$\E\l[\norm{U(\xbf;\o)}_{\ast}^{2} \r]\leq c\norm{A(\xbf)}_{\ast}^{2}$.
\end{assumption}
While \cref{assumption:random} is enough for obtaining the typical $\bigoh(1/\sqrt{T})$ rate in stochastic settings, \cref{assumption:relative} may allow us to recover the well-known order-optimal rate of $\bigoh(1/T)$ in deterministic settings. Intuitively, this improvement is explained by the fact that the noisy error measurements vanish while we approach a solution of \ref{eq:VI}. %
In~\cref{app:example}, we highlight random coordinate descent and random player updating as popular examples, which motivate \cref{assumption:relative}.

\section{Quantized generalized extra-gradient}
\label{sec:algo}
\subsection{System model and proposed algorithm}\label{sec:quant}

\begin{algorithm}[t]
\SetAlgoLined
    \KwIn{Local data, parameter vector (local copy) $X_t,Y_t$, learning rate 
    $\{\gamma_t\}$, and set of update steps $\Uc$}
\For{$t=1$ {\bfseries to} $T$}{
    \If{$t\in\Uc$}{
        \For{$i=1$ {\bfseries to} $K$}{
            Compute sufficient statistics and update quantization levels 
            $\bql_t$\;
        }
    }
    \For{$i=1$ {\bfseries to} $K$ }{ 
        Compute $V_{i,t}$, encode $c_{i,t}\leftarrow 
        \ENCODE\big(Q_{\bql_t}(V_{i,t});\bql_t\big)$, and broadcast $c_{i,t}$\par
        Receive $c_{i,t}$ from each processor $i$ and decode $\hat 
        V_{i,t}\leftarrow \DECODE(c_{i,t};\bql_t)$\par
        Aggregate $X_{t+1/2}\leftarrow X_{t}-\frac{\gamma_t}{K}\sum_{k=1}^K  \hat V_{k,t}$\par
        Compute $V_{i,t+1/2}$, encode $d_{i,t}\leftarrow 
        \ENCODE\big(Q_{\bql_t}(V_{i,t+1/2});\bql_t\big)$, and broadcast $d_{i,t}$\par
        Receive $d_{i,t}$ from each processor $i$ and decode $\hat 
        V_{i,t+1/2}\leftarrow \DECODE(d_{i,t};\bql_t)$\par
        Aggregate $Y_{t+1}\leftarrow Y_{t}-\frac{1}{K}\sum_{k=1}^K  \hat V_{k,t+1/2}$ and update $X_{t+1}\leftarrow\gamma_{t+1} Y_{t+1}$\par%
    }
}
\caption{\QEG: Loops are executed in parallel on processors. At certain steps, each processor computes sufficient statistics of a parametric distribution to estimate distribution of dual vectors.}
\label{QGEGalg}
\end{algorithm} We now describe the algorithmic framework unifying communication-efficient variants of generalized extra-gradient~(\EG) family of algorithms. In particular, we consider a synchronous and distributed system with $K$ processors along the lines of \eg data-parallel \sgd~\citep{QSGD,ALQ,NUQSGD,FL}. These processors can be cellphones and hospitals in FL or GPU resources in a data center. In multi-GPU systems, processors partition a large dataset among themselves such that each processor keeps only a local copy of the current parameter vector and has access to {\it independent} and {\it private  stochastic dual vectors}. At each iteration, each processor receives stochastic dual vectors from all other processors and aggregates them.
To accelerate training, stochastic dual vectors are first compressed by each processor before broadcasting to other peers and then decompressed before each aggregation step. %
We focus on unbiased compression where, {\it in expectation},  the  output of the decompression of a compressed vector is the same as the original uncompressed vector. We use $Q_{\bql_t}$ to denote a random and adaptive quantization function where the quantization levels $\bql_t$ may change over time. We use $V_{k,t}$ to denote the original (uncompressed) stochastic dual vector computed by process $k$ at time $t$.

Using multiple processors reduces computational costs significantly. However, communication costs to broadcast huge stochastic dual vectors is the main performance bottleneck in practice~\citep{QSGD}. In order to reduce communication costs and improve scalability, each processor receives and aggregates the compressed stochastic dual vectors from all peers to obtain the updated parameter vector.
Let $\bql_t=(\ql_0,\ql_1^t,\ldots,\ql_s^t,\ql_{s+1})$ denote the sequence of $s$ quantization levels {\it optimized} at iteration $t$ with  $0=\ql_0<\ql_1^t<\cdots<\ql_s^t<\ql_{s+1}=1$. We now define quantization function $Q_{\bql_t}$: 
\begin{definition}[Random quantization  function]\label{def:quantf}
Let $s\in\integers_+$ denote the number of quantization levels. Let $u\in[0,1]$ and $\bql_t=(\ql_0,\ql_1^t,\ldots,\ql_s^t,\ql_{s+1})$ denote the sequence of $s$ {\it quantization levels} at iteration $t$ with $0=\ql_0<\ql_1^t<\cdots<\ql_s^t<\ql_{s+1}=1$. Let ${\level{u}}$ denote the index of a level such that 
${\ql_{\level{u}}^t\leq u<\ql_{\level{u}+1}^t}$.
Let $\qcoeff_t(u)=(u-\ql_{\level{u}}^t)/(\ql_{\level{u}+1}^t-\ql_{\level{u}}^t)$ be the 
relative distance of $u$ to level $\level{u}+1$. We define the random function
$q_{\bql_t}(u):[0,1]\ra\{\ql_0,\ql_1^t,\ldots,\ql_s^t,\ql_{s+1}\}$ such that $q_{\bql_t}(u)=\ql_{\level{u}}^t$ with probability $1-\qcoeff_t(u)$ and 
$q_{\bql_t}(u)=\ql_{\level{u}+1}^t$ with probability $\qcoeff_t(u)$. Let $q\in\integers_+$ and  $\vbf\in\reals^d$. We define the random quantization of $\vbf$ as follows: 
\begin{align}\nn
 Q_{\bql_t}(\vbf):= \|\vbf\|_q\cdot \sbf\odot [q_{\bql_t}(u_1),\ldots,q_{\bql_t}(u_d)]^\top   
\end{align} where $\odot$ denotes the element-wise (Hadamard) product. 
\end{definition}

Let $\hat V_{k,t}=Q_{\bql_t}(V_{k,t})$ and $\hat V_{k,t+1/2}=Q_{\bql_t}(V_{k,t+1/2})$ denote the unbiased and quantized stochastic dual vectors for $k\in[K]$ and $t\in[T]$. We propose  {\it quantized} generalized extra-gradient (\QEG) family of algorithms with this update rule: 
\begin{align}
    \label{eq:QGEG}
    \tag{\QEG}
    \begin{split}
    X_{t+1 / 2}&=X_{t}-\frac{\gamma_t}{K}\sum_{k=1}^K  \hat V_{k,t} \\
    Y_{t+1}&=Y_{t}-\frac{1}{K}\sum_{k=1}^K\hat V_{k,t+1 / 2} \\
    X_{t+1}&=\gamma_{t+1} Y_{t+1}
    \end{split}
\end{align} where $(\hat V_{k,0},\hat V_{k,1},\ldots)$ and $(\hat V_{k,1/2},\hat V_{k,3/2},
\ldots)$ are the  sequences of {stochastic dual vectors} computed and quantized by processor $k\in[K]$.
Provided that $V_{k,t}$ and $V_{k,t+1/2}$ are stochastic dual vectors for $k\in[K]$, then $\frac{1}{K}\sum_{k=1}^K \hat V_{k,t}$ and $\frac{1}{K}\sum_{k=1}^K \hat V_{k,t+1/2}$ remain {\it unbiased} stochastic dual vectors.

\QEG is described in~\cref{QGEGalg}. 
In general, the decoded stochastic dual vectors are likely to be different from the original locally computed stochastic dual vectors. 

 A particularly appealing feature of  \eqref{eq:QGEG} formulation is that it enables us to unify  communication-efficient and distributed variants of a wide range of popular first-order methods for solving~\ref{eq:VI}s.
 In particular, one may observe that under different choices of $\hat V_{k,t}$ and $\hat V_{k,t+1/2}$,  {\it communication-efficient} variants of  stochastic dual averaging~\citep{Nes09}, stochastic dual extrapolation~\citep{Nes07}, and stochastic optimistic dual averaging~\citep{Pop80,RS13-NIPS,HAM21,HIMM22} in {\it multi-GPU settings} are special cases of \ref{eq:QGEG}:
 
 \begin{example}{\bf{Distributed stochastic dual averaging:}}
Consider the case where $\hat V_{k,t}\equiv0$ and $\hat V_{k,t+1/2}\equiv \hat g_{k,t+1/2}=Q_{\bql_t}(A(X_{t+1/2})+U_{k,t+1/2})$. This setting yields to $X_{t+1/2}=X_{t}$ and hence $\hat g_{k,t+1/2}=\hat g_{k,t}=\hat V_{k,t+1/2}$. Therefore, \ref{eq:QGEG} reduces to the communication-efficient stochastic dual averaging scheme:
\begin{align}
    \label{eq:DA}
    \tag{Quantized DA}
    \begin{split}
    Y_{t+1}&=Y_{t}-\frac{1}{K}\sum_{k=1}^K\hat g_{k,t}
    \\
    X_{t+1}&=\gamma_{t+1}Y_{t+1}
    \end{split}
\end{align}  
\end{example}  
\begin{example}{\bf{Distributed stochastic dual extrapolation}:}
Consider the case where $\hat V_{k,t}\equiv \hat g_{k,t}=Q_{\bql_t}(A(X_t)+U_{k,t})$ and $\hat V_{k,t+1/2}\equiv \hat g_{k,t+1/2}=Q_{\bql_t}(A(X_{t+1/2})+U_{k,t+1/2})$ are noisy oracle queries at $X_{t}$ and $X_{t+1/2}$, respectively.
Then \ref{eq:QGEG} provides the communication-efficient variant of the Nesterov's stochastic dual extrapolation method \citep{Nes07}:
\begin{align}
    \label{eq:DE}
    \tag{Quantized DE}
    \begin{split}
    X_{t+1 / 2}&=X_{t}-\frac{\g_t}{K}\sum_{k=1}^K  \hat g_{k,t}\\
    Y_{t+1}&=Y_{t}-\frac{1}{K}\sum_{k=1}^K\hat g_{k,t+1 / 2}\\
    X_{t+1}&=\gamma_{t+1} Y_{t+1}
    \end{split}
\end{align}
\end{example}
\begin{example}{\bf{Distributed stochastic optimistic dual averaging}:}
Consider the case $\hat V_{k,t}\equiv \hat g_{k,t-1/2}=Q_{\bql_t}(A(X_{t-1/2})+U_{k,t-1/2})$ and $\hat V_{k,t+1/2}\equiv\hat g_{k,t+1/2}=Q_{\bql_t}(A(X_{t+1/2})+U_{k,t+1/2})$ are the noisy oracle feedback at $X_{t-1/2}$ and $X_{t+1/2}$, respectively.
We then obtain the communication-efficient stochastic optimistic dual averaging method:
\begin{align}
    \label{eq:OptDA}
    \tag{Quantized OptDA}
    \begin{split}
    X_{t+1 / 2}&=X_{t}-\frac{\g_t}{K}\sum_{k=1}^K  \hat g_{k,t-1/2} \\
    Y_{t+1}&=Y_{t}-\frac{1}{K}\sum_{k=1}^K\hat g_{k,t+1 / 2} \\
    X_{t+1}&=\gamma_{t+1} Y_{t+1}
\end{split}
\end{align}
\end{example}

This general formulation of \ref{eq:QGEG} allows us to bring these variants under one umbrella and provide theoretical guarantees for all of them in a unified manner.

\subsection{Encoding}\label{sec:encoding}
To further reduce communication costs, we can apply information-theoretically coding schemes on top of quantization. Let $q\in\integers_+$. We first note that a vector $\vbf\in\reals^d$ can be {\it uniquely} represented by a tuple  $(\|\vbf\|_q,\signvec,\ubf)$
where $\|\vbf\|_q$ is the $L^q$ norm of $\vbf$,
$\signvec:=[\sign(v_1),\ldots,\sign(v_d)]^\top$ consists of signs of the coordinates $v_i$'s,
and $\ubf:=[u_1,\ldots,u_d]^\top$ with $u_i=|v_i|/\|\vbf\|_q$ are the normalized coordinates. Note that $0 \leq u_i\leq 1$ for all $i\in[d]$. The overall encoding, i.e., composition of coding and quantization, $\ENCODE(\|\vbf\|_q,\sbf,\qbf_{\bql_t}):\reals_+\times \{\pm 1\}^d\times \{\ql_0,\ql_1^t,\ldots,\ql_s^t,\ql_{s+1}\}^d\ra \{0,1\}^*$ in \cref{QGEGalg} uses a standard floating point encoding with $C_b$ bits to represent the positive scalar $\|\vbf\|_q$, encodes the sign of each coordinate with one bit, and finally applies an $integer$ encoding scheme $\Psi:\{\ql_0,\ql_1^t,\ldots,\ql_s^t,\ql_{s+1}\} \to \{0,1\}^*$ to {\it efficiently} encode each quantized and normalized coordinate $q_{\bql_t}(u_i)$ with the {\it minimum} expected code-length.
The overall decoding $\DECODE:\{0,1\}^*\to \reals^d$
 first reads $C_b$ bits to reconstruct $\|\vbf\|_q$. 
Then it applies $\Psi^{-1}:\{0,1\}^*\to\{\ql_0,\ql_1^t,\ldots,\ql_s^t,\ql_{s+1}\}$ to reconstruct normalized coordinates. The encoding/decoding details are provided in~\cref{app:encoding}.

\subsection{Adaptive quantization}\label{sec:aqsgd}
Instead of using a heuristically chosen sequence of quantization levels, adaptive quantization estimates distribution of uncompressed original vectors, i.e.,  dual vectors, by computing sufficient statistics of a parametric distribution, optimizes quantization levels to minimize the quantization error, and updates those levels adaptively throughout the course of training as the distribution changes.
Let $(\Omega_Q,\Fc_Q,\mathbb{P}_Q)$ denote a complete probability space. Let $\qbf_{\bql_t}\sim\mathbb{P}_Q$ represent $d$ variables sampled independently for random quantization in \cref{def:quantf}. %
Let $\vbf\in\reals^d$  denote a stochastic dual vector to be quantized. Given $\vbf$, we measure the quantization error by the variance of vector quantization, which is the trace of the covariance matrix: 
\begin{align}\label{var}
    \E_{\qbf_{\bql_t}}[\|Q_{\bql_t}(\vbf)-\vbf\|_2^2]=\|\vbf\|_q^2 \sum_{i=1}^d\sigma_Q^2(u_i;\bql_t)
\end{align} where $u_i=|v_i|/\|\vbf\|_q$ and $\sigma_Q^2(u;\bql_t)=\E_{\qbf_{\bql_t}}[(q_{\bql_t}(u)-u)^2]=(\ql_{\level{u}+1}^t-u)(u-\ql_{\level{u}}^t)$ is the variance  for a normalized  coordinate $u$. We optimize $\bql_t$ by minimizing the quantization variance:
\begin{align}\nn%
    \min_{\bql_t\in\feasl}\, \E_{\o}\E_{\qbf_{\bql_t}}\l[\|Q_{\bql_t}(g(\xbf_t;\o))-A(\xbf_t)\|_*^2\r]
\end{align}
where $\feasl=\{\bql:\ql_j\leq \ql_{j+1},~\forall~j,~\ql_0=0,~\ql_{s+1}=1 \}$ denotes the set of 
feasible solutions.

Since random quanitzation and random samples are statistically independent, we can solve the following equivalent problem: 
\begin{align}\tag{MinVar}\label{min_var}
    \min_{\bql_t\in\feasl}\, \E_{\o}\E_{\qbf_{\bql_t}}\l[\|Q_{\bql_t}(g(\xbf_t;\o))-g(\xbf_t;\o)\|_2^2\r]
\end{align}

To solve \ref{min_var}, we first sample $J$ stochastic dual vectors $\{g(\xbf_t;\o_1),\ldots,g(\xbf_t;\o_J)\}$\footnote{A more fine-grained analysis can be done by considering two sequences of adaptive levels one for $V_{k,t}$’s and another one for $V_{k,t+1/2}$’s. While in this paper, we quantize both sequences with the same quantization scheme, it is possible to further reduce quantization errors by considering two fine-grained quantization schemes at the cost of additional computations at processors.}. Let $\cdf_j(r)$ denote  the marginal cumulative distribution
function (CDF) of normalized coordinates conditioned on 
observing $\|g(\xbf_t;\o_j)\|_q$. By  the law of total expectation, \ref{min_var}  can be approximated by:
\begin{align}\tag{QAda}\label{Qadap}
\min_{\bql_t\in\feasl}
    \sum_{j=1}^J
        \|g(\xbf_t;\o_j)\|_q^2
            \sum_{i=0}^s
                \int_{\ql_i}^{\ql_{i+1}}
                   \sigma_Q^2(u;\bql_t)\ud \cdf_j(u)
    \equiv\min_{\bql_t\in\feasl}
        \sum_{i=0}^s
            \int_{\ql_i}^{\ql_{i+1}}
                \sigma_Q^2(u;\bql_t)\ud \tilde\cdf(u)
\end{align}
where $\tilde\cdf(u)= \sum_{j=1}^J\lambda_j\cdf_j(u)$ is the weighted sum of the conditional CDFs with $\lambda_j=\|g(\xbf_t;\o_j)\|_q^2/\sum_{j=1}^J\|g(\xbf_t;\o_j)\|_q^2$.

Finally, we solve~\ref{Qadap} efficiently by either updating levels one at a time or gradient descent along the lines of~\citep{ALQ}.

\section{Theoretical guarantees}
\label{sec:theory}

We first establish a variance error bound for a general unbiased and normalized compression scheme and a bound on the expected number of communication bits to encode $Q_\bql(\vbf)$, \ie the output of $\ENCODE$, which is introduced in~\cref{sec:encoding}. The detailed proofs are provided in the appendix.

Let $\ol\ql:=\max_{1\leq j\leq s} \ql_{j+1}/\ql_j$ and $d_{\mathrm{th}}=(2/\ql_1)^{\min\{q,2\}}$. Under general  $L^q$ normalization and sequence of $s$ quantization levels $\bql$, we first establish an upper bound on the variance of quantization:

\bth[Variance bound]\label{thm:varbound} 
Let $\vbf\in\reals^d$, $q\in\integers_+$, and $s\in\integers_+$. Let $\bql= (\ql_0,\ldots,\ql_{s+1})$ denote a sequence of $s$ {\it quantization levels} defined in~\cref{def:quantf}. The quantization of $\vbf$ in~\cref{def:quantf} is unbiased, \ie  $\E_{\qbf_\bql}[Q_\bql(\vbf)]=\vbf$. Furthermore, we have
\begin{align}\label[ineq]{varbound}
\E_{\qbf_\bql}[\|Q_\bql(\vbf)-\vbf\|_2^2]\leq\epsilon_Q\|\vbf\|_2^2,
\end{align} where
$\epsilon_Q=\frac{\ol\ql+\ol\ql^{-1}}{4}+\frac{1}{4}\ql_1^2d^{\frac{2}{\min\{q,2\}}}\one\{d\leq d_{\mathrm{th}}\}+\big(\ql_1d^{\frac{1}{\min\{q,2\}}}-1\big)\one\{d\geq d_{\mathrm{th}}\}-\frac{1}{2}$ and $\one$ is the indicator function. 
\eth

\cref{thm:varbound} implies that if $g(\xbf;\o)$ is an unbiased stochastic dual vector  with a bounded absolute variance  $\sigma^2$, then $Q_\bql(g(\xbf;\o))$ will be an unbiased stochastic dual vector with a variance upper bound $\epsilon_Q\sigma^2$. Note that, the dominant term $\frac{1}{4}\ql_1^2d^{\frac{2}{\min\{q,2\}}}\one\{d\leq d_{\mathrm{th}}\}+\big(\ql_1d^{\frac{1}{\min\{q,2\}}}-1\big)\one\{d\geq d_{\mathrm{th}}\}$ monotonically decreases as the number of quantization levels increases.  Unlike~\citep[Theorem 3.2]{QSGD} and~\citep[Theorem 4]{NUQSGD} that hold under the special cases of $L^2$ normalization with uniform and exponentially spaced levels, respectively, our bound in~\cref{thm:varbound} holds under general $L^q$ normalization and arbitrary sequence of quantization levels. For the special case of $L^2$ normalization in the regime of large $d$, which is the case in practice, our bound in~\cref{thm:varbound} is ${\Oc} (\ell_1\sqrt{d})$, which is arbitrarily smaller than ${\Oc} (\sqrt{d}/s)$ and ${\Oc} (2^{-s}\sqrt{d})$ in \citep[Theorem 3.2]{QSGD} and \citep[Theorem 4]{NUQSGD}, respectively, because $\ell_1$ is adaptively designed to minimize the variance of quantization. Furthermore, unlike the bound in \citep[Theorem 2]{ALQ} that requires an inner optimization problem over an auxiliary parameter $p$, the bound in~\cref{thm:varbound} is provided in an explicit form without an inner optimization problem, which matches the known $\Omega(\sqrt{d})$ lower bound.

\bth[Code-length bound]\label{thm:codebound}
Let $p_j$ denote the probability of occurrence of $\ql_j$ (weight of symbol $\ql_j$) for $j\in[s]$.
Under the setting specified in~\cref{thm:varbound}, the expectation $\E_{\o}\E_{\qbf_\bql}[|\ENCODE(Q_\bql(g(\xbf;\o));\bql)|]$ of the number of bits to encode $Q_\bql(g(\xbf;\o))$ is bounded by
\begin{align}\label{codebound}%
\begin{split}
\E_{\o}\E_{\qbf_\bql}[|\ENCODE(Q_\bql(g(\xbf;\o));\bql)|]&= \bigoh\Big((\sum_{j=1}^sp_j\log(1/p_j)-p_0)d\Big).%
\end{split}
\end{align} %
\eth
Note $\{p_0,\ldots,p_{s+1}\}$ can be computed efficiently using the weighted sum of the conditional CDFs in~\ref{Qadap} and quantization levels. We provide their expressions in~\cref{app:pr_code}. %
Unlike~\citep[Theorem 3.4]{QSGD} and ~\citep[Theorem 5]{NUQSGD} that hold under the special cases of $L^2$ normalization, our bound in~\cref{thm:codebound} holds under general $L^q$ normalization.  For the special case of $L^2$ normalization with $s=\sqrt{d}$ as in~\citep[Theorem 3.4]{QSGD}, our bound in~\cref{thm:codebound} can be arbitrarily smaller than~\citep[Theorem 3.4]{QSGD} and ~\citep[Theorem 5]{NUQSGD} depending on $\{p_0,\ldots,p_{s+1}\}$. Compared to \citep[Theorem 3]{ALQ}, our bound in~\cref{thm:codebound} does not have an additional $n_{\ell_1,d}$ term. We show that a total expected number of $\bigoh(Kd/\epsilon)$ bits are required to reach an $\epsilon$ gap, which matches the lower bound developed for convex optimization problems with finite-sum structures~\citep{tsitsiklis1987communication,korhonen2021towards}.

We finally present the convergence guarantees for \ref{eq:QGEG} given access to stochastic dual vectors under both {\it absolute noise} and {\it relative noise} models in \cref{assumption:random,assumption:relative}, respectively.

\bth[\ref{eq:QGEG} under absolute noise]\label{thm:convQabs}
Let $\Cc\subset\reals^d$ denote a compact neighborhood of a solution for \eqref{eq:VI} and let %
$D^2:=\sup_{X\in\Cc} \|X-X_0\|^2$. Suppose that the oracle and the problem~\eqref{eq:VI} satisfy Assumptions~\ref{assumption:ex} and~\ref{assumption:random}, respectively,~\cref{QGEGalg}  
is executed for $T$ iterations on $K$ processors with an adaptive step-size $\gamma_t=K(1+\sum_{i=1}^{t-1}\sum_{k=1}^K\|\hat V_{k,i}-\hat V_{k,i+1/2}\|^2)^{-1/2}$, and quantization levels are updated $J$ times where $\bql_j$ with variance bound $\epsilon_{Q,j}$ in \eqref{varbound} and code-length bound $N_{Q,j}$ in \eqref{codebound} is used for $T_j$ iterations with $\sum_{j=1}^J T_j=T$. Then we have
\begin{align}
\E\Big[\gap_{\Cc}\Big(\frac{1}{T}\sum_{t=1}^T X_{t+1/2}\Big)\Big]=\bigoh\Big(\frac{(\sum_{j=1}^J \sqrt{\epsilon_{Q,j}T_j/T}M+\sigma) D^2}{\sqrt{TK}}\Big)\nn.
\end{align} In addition,~\cref{QGEGalg} requires each processor to send at most $ \frac{2}{T}\sum_{j=1}^J{T_j N_{Q,j}}$ communication bits per iteration in expectation. 
\eth

We now establish {\it fast rate} of $\Oc (1/T)$ under {\it relative noise} and a mild regularity condition: 

\begin{assumption}[Co-coercivity]\label{assumption:co}
Let $\smooth >0$. We assume that operator $A$ is $\smooth$-cocoercive:
\begin{equation}
\label{eq:coco}
    \braket{A(\xbf)-A(\xbf')}{\xbf-\xbf'}\geq \smooth \dnorm{A(\xbf)-A(\xbf')}^{2}\;\;\text{for all}\;\;\xbf,\xbf'\in \reals^{d}.
\end{equation}
\end{assumption} 
For a panoramic view of this class of operators, we refer the reader to~\citep{BC17}.

\begin{remark}The order-optimal rate of $\Oc (1/\sqrt{T})$ under absolute noise does not require co-cocercivity. Our adaptive step-size also does not depend on the noise model or co-coercivity. Co-coercivity is required only to achieve {\it fast rate} of $\Oc (1/T)$ in the case of {\it relative noise}.
\end{remark}

\bth[\ref{eq:QGEG} under relative noise]\label{thm:convQrel}
Let $\Cc\subset\reals^d$ denote a compact neighborhood of a solution for \eqref{eq:VI} and let %
$D^2:=\sup_{X\in\Cc} \|X-X_0\|^2$. Suppose that the oracle and the problem~\eqref{eq:VI} satisfy Assumptions~\ref{assumption:ex},~\ref{assumption:relative}, and \ref{assumption:co},~\cref{QGEGalg}  
is executed for $T$ iterations on $K$ processors with an adaptive step-size $\gamma_t=K(1+\sum_{i=1}^{t-1}\sum_{k=1}^K\|\hat V_{k,i}-\hat V_{k,i+1/2}\|^2)^{-1/2}$,
and quantization levels are updated $J$ times where $\bql_j$ with variance bound $\epsilon_{Q,j}$ in \eqref{varbound} and code-length bound $N_{Q,j}$ in \eqref{codebound} is used for $T_j$ iterations with $\sum_{j=1}^J T_j=T$. Then we have
\begin{align}
\E\Big[\gap_{\Cc}\Big(\frac{1}{T}\sum_{t=1}^T X_{t+1/2}\Big)\Big]=\bigoh\Big(\frac{\big((c+1)\sum_{j=1}^JT_j \epsilon_{Q,j}/T+c\big) D^2}{KT}\Big).\nn
\end{align} In addition,~\cref{QGEGalg} requires each processor to send at most $ \frac{2}{T}\sum_{j=1}^J{T_j N_{Q,j}}$ communication bits per iteration in expectation. 
\eth 

To the best of our knowledge, our results in~\cref{thm:convQabs,thm:convQrel} are the first ones proving that increasing the number of processors accelerates convergence for general monotone \ref{eq:VI}s under an adaptive step size. \cref{thm:convQabs,thm:convQrel} show that we can attain a fast rate of  $\Oc(1/T)$ and an order-optimal $\Oc(1/\sqrt{T})$ without prior knowledge on the noise profile while significantly reducing communication costs.

Compared to saddle point problems, our rates are optimal. This can be verified by lower bounds in~\citep{beznosikov2020distributed}. For convex problems in deterministic settings, the rate can be improved to $\Oc(1/T^2)$ via acceleration. However, it is known that in the stochastic and distributed settings, our rates cannot be improved even with acceleration. E.g., for convex and smooth problems under absolute noise model, the lower bound of $\Omega (\frac{1}{\sqrt{TK}})$ can be established by~\citep[Theorem 1]{woodworth2021min} and setting the number of gradients per round to one.

In~\cref{app:tradeoff}, we build on \cref{thm:convQabs,thm:convQrel} to capture the trade-off between the number of iterations to converge and time per iteration, which includes total time required to update a model on each GPU.

\section{Experimental evaluation}
\label{sec:exp}
\begin{figure*}[t]
\centering
\begin{minipage}{0.3\textwidth}
\centering
    \includegraphics[width=\textwidth]{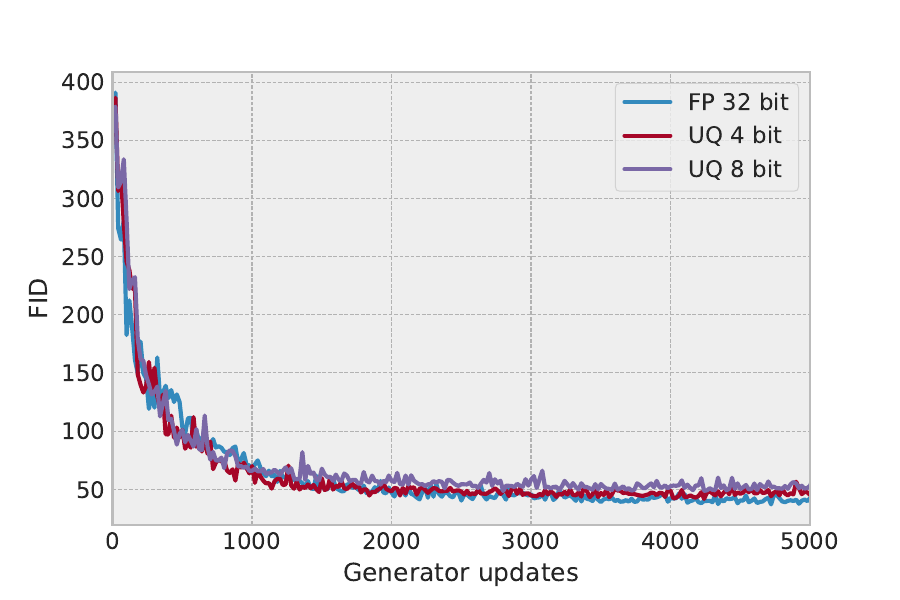}
    \label{fig:fid_vs_time}
\end{minipage}
\begin{minipage}{0.3\textwidth}
\centering
    \includegraphics[width=\textwidth]{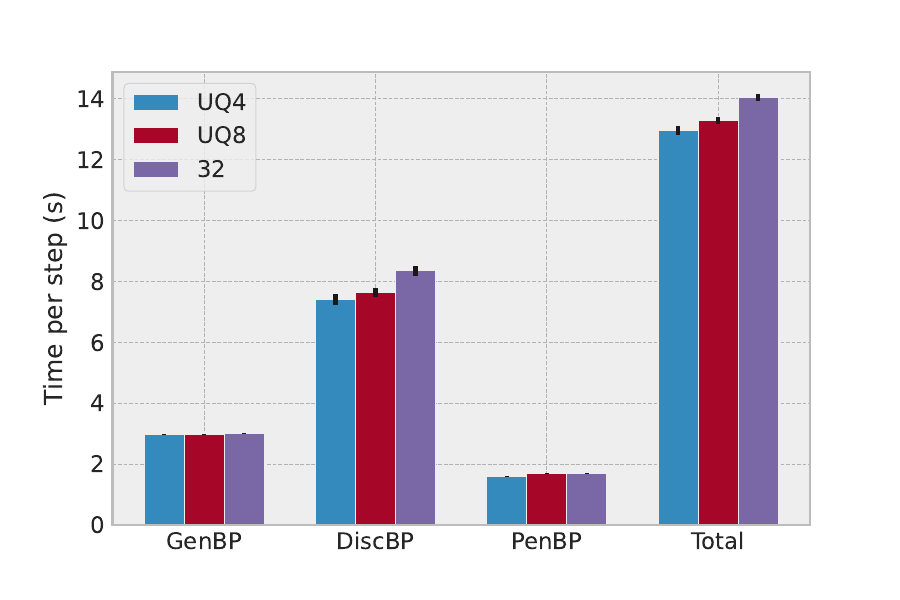}
    \label{fig:fine_grained}
\end{minipage}%
\begin{minipage}{0.36\textwidth}
\centering
\captionsetup{type=table} %
\begin{tabular}{p{0.5cm}p{0.7cm}p{0.7cm}p{0.7cm}p{0.5cm}}
\toprule
Mode &     GenBP &    DiscBP &     PenBP &      Total \\
\midrule
UQ4 & 2.99  & 7.40 & 1.59 & 12.96 \\
UQ8 & 2.99  & 7.65  & 1.69  & 13.29 \\
FP32 & 3.00  & 8.36  & 1.69  & 14.05 \\
\bottomrule
\end{tabular}
\label{tab:fine_grained}
\end{minipage}
    \caption{FID evolution during training  (left). We compare full-precision ExtraAdam with a simple instantiation of Q-GenX. FID stands for Frechet inception distance, which is a standard GAN quality metric introduced in~\citep{FID}. Fine grained comparison of average \texttt{.backward()} times  on generator, discriminator, gradient penalty as well as total training time (s)  (middle and right). The \texttt{.backward()} function is where pytorch DistributedDataParallel (DDP) handles gradient exchange.}
    \label{fine_grained}
\end{figure*}

In order to validate our theoretical results, we build on the code base of \citet{GBVV+19} and run an instantiation of Q-GenX obtained by combining ExtraAdam with the compression offered by the \texttt{torch\_cgx} pytorch extension of~\citet{CGX}, and train a WGAN-GP~\citep{arjovsky2017wasserstein} on CIFAR10~\citep{CIFAR10}.

Since \texttt{torch\_cgx} uses OpenMPI~\citep{gabriel04:_open_mpi} as its communication backend, we use OpenMPI  as the communication backend for the full gradient as well for a fairer comparison.
We deliberately do \emph{not} tune any hyperparameters to fit the larger batchsize since simliar to~\citep{GBVV+19}, we do not claim to set a new SOTA with these experiments but simply want to show that our theory holds up in practice and can potentially lead to improvements. For this, we present a basic experiment showing that even for a {\it very small problem size} and  {\it a heuristic base compression method of cgx}, we can achieve a {\it noticeable speedup} of around $8\%$. We expect further gains to be achievable for larger problems and more advanced compression methods. Given differences in terms of settings and the {\it lack of any code, let alone an efficient implementations that can be used in a real-world setting (i.e. CUDA kernels integrated with networking)}, it is difficult to impossible to conduct a fair comparison with \citet{beznosikov2021distributed}. More details and a comparison with QSGDA of \citet{beznosikov2022unifiedtheory} are provided in~\cref{app:experiment}. We are not aware of any other existing method dealing with the same problem in our paper without the cost associated with variance reduction.%

We follow exactly the setup of \citep{GBVV+19} except that we share an effective batch size of $1024$ across 3 nodes (strong scaling) connected via Ethernet, and use Layernorm~\citep{ba2016layer} instead of Batchnorm~\citep{ioffe2015batch} since Batchnorm is known to be challenging to work with in distributed training as well as interacting badly with the WGAN-GP penalty. The results are shown in \Cref{fine_grained} (left) showing evolution of FID. We note that we do \emph{not} scale the learning rate or any other hyperparameters to account for these two changes so this experiment is \emph{not} meant to claim SOTA performance, merely to illustrate that 
\begin{enumerate}
    \item Even with the simplest possible unbiased quantization on a relatively small-scale setup, we can observe a speedup (about $8\%$).
    \item This speedup does not drastically change the performance.
\end{enumerate}
We compare training using the full gradient  of 32 bit (FP32) to training with gradients compressed to $8$ (UQ8) and $4$ bits  (UQ4) using a bucket size of $1024$. \Cref{fine_grained} (middle and right) shows a more fine grained breakdown of the time used for back propagation (BP)  where the network activity takes place. GenBP, DiscBP and PenBP refer to the backpropagation for generator, discriminator, and the calculation of the gradient penalty, respectively. Total refers to the sum of these times.

\section{Conclusions }%
\label{sec:conc}
We have considered mononote \ref{eq:VI}s in a synchronous and multi-GPU  setting where multiple processors compute independent and private stochastic dual vectors in parallel. We proposed \QEG, which employs unbiased and adaptive compression methods tailored to a generic unifying framework for solving \ref{eq:VI}s. Without knowing the noise profile in advance, we have  obtained an adaptive step-size rule, which achieves a fast rate of  $\Oc(1/T)$ under relative noise, and an order-optimal $\Oc(1/\sqrt{T})$ in the absolute noise case along with improved guarantees on the expected number of communication bits. Our results show that increasing the number of processors accelerates convergence.

Developing new  \ref{eq:VI}-solvers for {\it asynchronous} settings and establishing convergence guarantees while relaxing co-coercivity assumption on the operator are interesting problems left for future work. Expected co-coercivity has been used as a more relaxed noise model~\citep{loizou2021stochastic}. It is interesting to study monotone~\ref{eq:VI}s under expected co-coercivity and adaptive step-sizes in the future.

\subsubsection*{Acknowledgments}
The authors would like to thank Fartash Faghri, Yang Linyan, Ilia Markov, and Hamidreza Ramezanikebrya for
helpful discussions. This project has received funding from the European Research Council (ERC) under the European Union’s Horizon 2020 research and innovation programme (grant agreement n° 725594 - time-data). This work was supported by the Swiss National Science Foundation (SNSF) under grant number 200021\_205011.

The work of Ali Ramezani-Kebrya was in part supported by the Research Council of Norway, through its Centre for Research-based Innovation funding scheme (Visual Intelligence under grant no. 309439),
and Consortium Partners.
\doclicenseThis

\bibliographystyle{iclr2023_conference}
\bibliography{bibtex/Ref,bibtex/Bibliography-PM}

\begin{thebibliography}{76}
\providecommand{\natexlab}[1]{#1}
\providecommand{\url}[1]{\texttt{#1}}
\expandafter\ifx\csname urlstyle\endcsname\relax
  \providecommand{\doi}[1]{doi: #1}\else
  \providecommand{\doi}{doi: \begingroup \urlstyle{rm}\Url}\fi

\bibitem[Abadi et~al.(2016)Abadi, Barham, Chen, Davis, Dean, Devin, Ghemawat,
  Irving, Isard, Kudlur, Levenberg, Monga, Moore, Murray, Steiner, Tucker,
  Vasudevan, Warden, Wicke, Yu, and Zheng]{Abadi}
Mart\'{i}n Abadi, Paul Barham, Zhifeng Chen, Andy Davis, Jeffrey Dean, Matthieu
  Devin, Sanjay Ghemawat, Geoffrey Irving, Michael Isard, Manjunath Kudlur,
  Josh Levenberg, Rajat Monga, Sherry Moore, Derek~G. Murray, Benoit Steiner,
  Paul Tucker, Vijay Vasudevan, Pete Warden, Martin Wicke, Yuan Yu, and
  Xiaoqiang Zheng.
\newblock {TensorFlow}: A system for large-scale machine learning.
\newblock In \emph{USENIX Symposium on Operating Systems Design and
  Implementation (OSDI)}, 2016.

\bibitem[Agarwal et~al.(2021)Agarwal, Wang, Lee, Venkataraman, and
  Papailiopoulos]{agarwal2021adaptive}
Saurabh Agarwal, Hongyi Wang, Kangwook Lee, Shivaram Venkataraman, and Dimitris
  Papailiopoulos.
\newblock Adaptive gradient communication via critical learning regime
  identification.
\newblock In \emph{Proceedings of Machine Learning and Systems (MLSys)}, 2021.

\bibitem[Alistarh et~al.(2017)Alistarh, Grubic, Li, Tomioka, and
  Vojnovic]{QSGD}
Dan Alistarh, Demjan Grubic, Jerry~Z. Li, Ryota Tomioka, and Milan Vojnovic.
\newblock {QSGD}: Communication-efficient {SGD} via gradient quantization and
  encoding.
\newblock In \emph{Advances in Neural Information Processing Systems
  (NeurIPS)}, 2017.

\bibitem[Antonakopoulos \& Mertikopoulos(2021)Antonakopoulos and
  Mertikopoulos]{AM21}
Kimon Antonakopoulos and Panayotis Mertikopoulos.
\newblock Adaptive first-order methods revisited: {Convex} optimization without
  {Lipschitz} requirements.
\newblock In \emph{Advances in Neural Information Processing Systems
  (NeurIPS)}, 2021.

\bibitem[Antonakopoulos et~al.(2019)Antonakopoulos, Belmega, and
  Mertikopoulos]{ABM19}
Kimon Antonakopoulos, Veronica Belmega, and Panayotis Mertikopoulos.
\newblock An adaptive mirror-prox algorithm for variational inequalities with
  singular operators.
\newblock In \emph{Advances in Neural Information Processing Systems
  (NeurIPS)}, 2019.

\bibitem[Antonakopoulos et~al.(2021)Antonakopoulos, Pethick, Kavis,
  Mertikopoulos, and Cevher]{Universal}
Kimon Antonakopoulos, Thomas Pethick, Ali Kavis, Panayotis Mertikopoulos, and
  Volkan Cevher.
\newblock Sifting through the noise: {Universal} first-order methods for
  stochastic variational inequalities.
\newblock In \emph{Advances in Neural Information Processing Systems
  (NeurIPS)}, 2021.

\bibitem[Arjovsky et~al.(2017)Arjovsky, Chintala, and
  Bottou]{arjovsky2017wasserstein}
Martin Arjovsky, Soumith Chintala, and L{\'e}on Bottou.
\newblock Wasserstein generative adversarial networks.
\newblock In \emph{International Conference on Machine Learning (ICML)}, 2017.

\bibitem[Ba et~al.(2016)Ba, Kiros, and Hinton]{ba2016layer}
Jimmy~Lei Ba, Jamie~Ryan Kiros, and Geoffrey~E. Hinton.
\newblock Layer normalization.
\newblock \emph{arXiv preprint arXiv:1607.06450}, 2016.

\bibitem[Bach \& Levy(2019)Bach and Levy]{BL19}
Francis Bach and Kfir~Yehuda Levy.
\newblock A universal algorithm for variational inequalities adaptive to
  smoothness and noise.
\newblock In \emph{Conference on Learning Theory (COLT)}, 2019.

\bibitem[Bauschke \& Combettes(2017)Bauschke and Combettes]{BC17}
Heinz~H. Bauschke and Patrick~L. Combettes.
\newblock \emph{Convex Analysis and Monotone Operator Theory in Hilbert
  Spaces}.
\newblock Springer, 2017.

\bibitem[Bekkerman et~al.(2011)Bekkerman, Bilenko, and Langford]{Scaleup}
Ron Bekkerman, Mikhail Bilenko, and John Langford.
\newblock \emph{Scaling up Machine Learning: Parallel and Distributed
  Approaches}.
\newblock Cambridge University Press, 2011.

\bibitem[Bernstein et~al.(2018)Bernstein, Wang, Azizzadenesheli, and
  Anandkumar]{signSGD}
Jeremy Bernstein, Yu-Xiang Wang, Kamyar Azizzadenesheli, and Anima Anandkumar.
\newblock sign{SGD}: Compressed optimisation for non-convex problems.
\newblock In \emph{International Conference on Machine Learning (ICML)}, 2018.

\bibitem[Beznosikov et~al.(2020)Beznosikov, Samokhin, and
  Gasnikov]{beznosikov2020distributed}
Aleksandr Beznosikov, Valentin Samokhin, and Alexander Gasnikov.
\newblock Distributed saddle-point problems: Lower bounds, optimal and robust
  algorithms.
\newblock \emph{arXiv preprint arXiv:2010.13112}, 2020.

\bibitem[Beznosikov et~al.(2021)Beznosikov, Richt{\'a}rik, Diskin, Ryabinin,
  and Gasnikov]{beznosikov2021distributed}
Aleksandr Beznosikov, Peter Richt{\'a}rik, Michael Diskin, Max Ryabinin, and
  Alexander Gasnikov.
\newblock Distributed methods with compressed communication for solving
  variational inequalities, with theoretical guarantees.
\newblock \emph{arXiv:2110.03313}, 2021.

\bibitem[Beznosikov et~al.(2022)Beznosikov, Gorbunov, Berard, and
  Loizou]{beznosikov2022unifiedtheory}
Aleksandr Beznosikov, Eduard Gorbunov, Hugo Berard, and Nicolas Loizou.
\newblock Stochastic gradient descent-ascent: Unified theory and new efficient
  methods.
\newblock \emph{arXiv preprint arXiv:2202.07262}, 2022.

\bibitem[Biewald(2020)]{wandb}
Lukas Biewald.
\newblock Experiment tracking with weights and biases, 2020.
\newblock URL \url{https://www.wandb.com/}.
\newblock Software available from wandb.com.

\bibitem[Chilimbi et~al.(2014)Chilimbi, Suzue, Apacible, and
  Kalyanaraman]{Projectadam}
Trishul Chilimbi, Yutaka Suzue, Johnson Apacible, and Karthik Kalyanaraman.
\newblock Project {Adam}: Building an efficient and scalable deep learning
  training system.
\newblock In \emph{USENIX Symposium on Operating Systems Design and
  Implementation (OSDI)}, 2014.

\bibitem[Coates et~al.(2013)Coates, Huval, Wang, Wu, Catanzaro, and Ng]{Coates}
Adam Coates, Brody Huval, Tao Wang, David Wu, Bryan Catanzaro, and Andrew~Y.
  Ng.
\newblock Deep learning with {COTS HPC} systems.
\newblock In \emph{International Conference on Machine Learning (ICML)}, 2013.

\bibitem[Cover \& Thomas(2006)Cover and Thomas]{InfTheory}
Thomas~M. Cover and Joy~A. Thomas.
\newblock \emph{Elements of Information Theory}.
\newblock WILEY, 2006.

\bibitem[Cummiskey et~al.(1973)Cummiskey, Jayant, and Flanagan]{Cummiskey}
P.~Cummiskey, Nikil~S. Jayant, and James~L. Flanagan.
\newblock Adaptive quantization in differential {PCM} coding of speech.
\newblock \emph{Bell System Technical Journal}, 52(7):\penalty0 1105--1118,
  1973.

\bibitem[Davies et~al.(2021)Davies, Gurunanthan, Moshrefi, Ashkboos, and
  Alistarh]{lattice}
Peter Davies, Vijaykrishna Gurunanthan, Niusha Moshrefi, Saleh Ashkboos, and
  Dan Alistarh.
\newblock New bounds for distributed mean estimation and variance reduction.
\newblock In \emph{International Conference on Learning Representations
  (ICLR)}, 2021.

\bibitem[Dean et~al.(2012)Dean, Corrado, Monga, Chen, Devin, Mao, Ranzato,
  Senior, Tucker, Yang, Le, and Ng]{Dean12}
Jeffrey Dean, Greg~S. Corrado, Rajat Monga, Kai Chen, Matthieu Devin, Mark~Z.
  Mao, Marc'aurelio Ranzato, Andrew Senior, Paul Tucker, Ke~Yang, Quoc Le, and
  Andrew~Y. Ng.
\newblock Large scale distributed deep networks.
\newblock In \emph{Advances in Neural Information Processing Systems
  (NeurIPS)}, 2012.

\bibitem[Duchi et~al.(2015)Duchi, Chaturapruek, and R\'{e}]{Duchi15}
John~C. Duchi, Sorathan Chaturapruek, and Christopher R\'{e}.
\newblock Asynchronous stochastic convex optimization.
\newblock In \emph{Advances in Neural Information Processing Systems
  (NeurIPS)}, 2015.

\bibitem[Elias(1975)]{Elias}
Peter Elias.
\newblock Universal codeword sets and representations of the integers.
\newblock \emph{{IEEE} Transactions on Information Theory}, 21(2):\penalty0
  194--203, 1975.

\bibitem[Facchinei \& Pang(2003)Facchinei and Pang]{facchinei2003finite}
Francisco Facchinei and Jong-Shi Pang.
\newblock \emph{Finite-dimensional variational inequalities and complementarity
  problems}.
\newblock Springer, 2003.

\bibitem[Faghri et~al.(2020)Faghri, Tabrizian, Markov, Alistarh, Roy, and
  Ramezani-Kebrya]{ALQ}
Fartash Faghri, Iman Tabrizian, Ilia Markov, Dan Alistarh, Daniel~M. Roy, and
  Ali Ramezani-Kebrya.
\newblock Adaptive gradient quantization for data-parallel {SGD}.
\newblock In \emph{Advances in Neural Information Processing Systems
  (NeurIPS)}, 2020.

\bibitem[Gabriel et~al.(2004)Gabriel, Fagg, Bosilca, Angskun, Dongarra,
  Squyres, Sahay, Kambadur, Barrett, Lumsdaine, Castain, Daniel, Graham, and
  Woodall]{gabriel04:_open_mpi}
Edgar Gabriel, Graham~E. Fagg, George Bosilca, Thara Angskun, Jack~J. Dongarra,
  Jeffrey~M. Squyres, Vishal Sahay, Prabhanjan Kambadur, Brian Barrett, Andrew
  Lumsdaine, Ralph~H. Castain, David~J. Daniel, Richard~L. Graham, and
  Timothy~S. Woodall.
\newblock Open {MPI}: Goals, concept, and design of a next generation {MPI}
  implementation.
\newblock In \emph{Proceedings, 11th European PVM/MPI Users' Group Meeting},
  pp.\  97--104, Budapest, Hungary, September 2004.

\bibitem[Gidel et~al.(2019)Gidel, Berard, Vignoud, Vincent, and
  Lacoste-Julien]{GBVV+19}
Gauthier Gidel, Hugo Berard, Ga{\"e}tan Vignoud, Pascal Vincent, and Simon
  Lacoste-Julien.
\newblock A variational inequality perspective on generative adversarial
  networks.
\newblock In \emph{International Conference on Learning Representations
  (ICLR)}, 2019.

\bibitem[Goodfellow et~al.(2020)Goodfellow, Pouget-Abadie, Mirza, Xu,
  Warde-Farley, Ozair, Courville, and Bengio]{goodfellow2020generative}
Ian Goodfellow, Jean Pouget-Abadie, Mehdi Mirza, Bing Xu, David Warde-Farley,
  Sherjil Ozair, Aaron Courville, and Yoshua Bengio.
\newblock Generative adversarial networks.
\newblock \emph{Communications of the ACM}, 63\penalty0 (11):\penalty0
  139--144, 2020.

\bibitem[Guo et~al.(2020)Guo, Liu, Wang, Han, Li, Lu, and
  Hu]{guo2020accelerating}
Jinrong Guo, Wantao Liu, Wang Wang, Jizhong Han, Ruixuan Li, Yijun Lu, and
  Songlin Hu.
\newblock Accelerating distributed deep learning by adaptive gradient
  quantization.
\newblock In \emph{IEEE International Conference on Acoustics, Speech and
  Signal Processing (ICASSP)}, 2020.

\bibitem[Gupta et~al.(2015)Gupta, Agrawal, Gopalakrishnan, and
  Narayanan]{Gupta}
Suyog Gupta, Ankur Agrawal, Kailash Gopalakrishnan, and Pritish Narayanan.
\newblock Deep learning with limited numerical precision.
\newblock In \emph{International Conference on Machine Learning (ICML)}, 2015.

\bibitem[Heusel et~al.(2017)Heusel, Ramsauer, Unterthiner, Nessler, and
  Hochreiter]{FID}
Martin Heusel, Hubert Ramsauer, Thomas Unterthiner, Bernhard Nessler, and Sepp
  Hochreiter.
\newblock {GANs} trained by a two time-scale update rule converge to a local
  nash equilibrium.
\newblock In \emph{Advances in Neural Information Processing Systems
  (NeurIPS)}, 2017.

\bibitem[Hsieh et~al.(2021)Hsieh, Antonakopoulos, and Mertikopoulos]{HAM21}
Yu-Guan Hsieh, Kimon Antonakopoulos, and Panayotis Mertikopoulos.
\newblock Adaptive learning in continuous games: {Optimal} regret bounds and
  convergence to {Nash} equilibrium.
\newblock In \emph{Conference on Learning Theory (COLT)}, 2021.

\bibitem[Hsieh et~al.(2022)Hsieh, Iutzeler, Malick, and Mertikopoulos]{HIMM22}
Yu-Guan Hsieh, Franck Iutzeler, J{\'e}r{\^o}me Malick, and Panayotis
  Mertikopoulos.
\newblock Multi-agent online optimization with delays: {Asynchronicity},
  adaptivity, and optimism.
\newblock \emph{Journal of Machine Learning Research}, 23:\penalty0 1--49,
  2022.

\bibitem[Ioffe \& Szegedy(2015)Ioffe and Szegedy]{ioffe2015batch}
Sergey Ioffe and Christian Szegedy.
\newblock Batch normalization: Accelerating deep network training by reducing
  internal covariate shift.
\newblock In \emph{International Conference on Machine Learning (ICML)}, 2015.

\bibitem[Juditsky et~al.(2011)Juditsky, Nemirovski, and Tauvel]{JNT11}
Anatoli Juditsky, Arkadi~Semen Nemirovski, and Claire Tauvel.
\newblock Solving variational inequalities with stochastic mirror-prox
  algorithm.
\newblock \emph{Stochastic Systems}, 1\penalty0 (1):\penalty0 17--58, 2011.

\bibitem[Kairouz et~al.(2021)Kairouz, McMahan, Avent, Bellet, Bennis, Bhagoji,
  Bonawitz, Charles, Cormode, Cummings, D'Oliveira, Rouayheb, Evans, Gardner,
  Garrett, Gasc\'{o}n, Ghazi, Gibbons, Gruteser, Harchaoui, He, He, Huo,
  Hutchinson, Hsu, Jaggi, Javidi, Joshi, Khodak, Kone\u{c}n\'{y}, Korolova,
  Koushanfar, Koyejo, Lepoint, Liu, P., Mohri, Nock, \"{O}zg\"{u}r, Pagh,
  Raykova, Qi, Ramage, Raskar, Song, Song, Stich, Sun, Suresh, Tram\`{e}r,
  Vepakomma, Wang, Xiong, Xu, Yang, Yu, Yu, and Zhao]{FL}
P.~Kairouz, H.~B. McMahan, B.~Avent, A.~Bellet, M.~Bennis, A.~N. Bhagoji,
  K.~Bonawitz, Z.~Charles, G.~Cormode, R.~Cummings, R.~G.~L. D'Oliveira, S.~E.
  Rouayheb, D.~Evans, J.~Gardner, Z.~Garrett, A.~Gasc\'{o}n, B.~Ghazi, P.~B.
  Gibbons, M.~Gruteser, Z.~Harchaoui, C.~He, L.~He, Z.~Huo, B.~Hutchinson,
  J.~Hsu, M.~Jaggi, T.~Javidi, G.~Joshi, M.~Khodak, J.~Kone\u{c}n\'{y},
  A.~Korolova, F.~Koushanfar, S.~Koyejo, T.~Lepoint, Y.~Liu, P., M.~Mohri,
  R.~Nock, A.~\"{O}zg\"{u}r, R.~Pagh, M.~Raykova, H.~Qi, D.~Ramage, R.~Raskar,
  D.~Song, W.~Song, S.~U. Stich, Z.~Sun, A.~T. Suresh, F.~Tram\`{e}r,
  P.~Vepakomma, J.~Wang, L.~Xiong, Z.~Xu, Q.~Yang, F.~X. Yu, H.~Yu, and
  S.~Zhao.
\newblock Advances and open problems in federated learning.
\newblock \emph{Foundations and Trends\textsuperscript{\textregistered} in
  Machine Learning}, 14\penalty0 (1–2):\penalty0 1--210, 2021.

\bibitem[Kavis et~al.(2019)Kavis, Levy, Bach, and Cevher]{KLBC19}
Ali Kavis, Kfir~Yehuda Levy, Francis Bach, and Volkan Cevher.
\newblock {UnixGrad}: {A} universal, adaptive algorithm with optimal guarantees
  for constrained optimization.
\newblock In \emph{Advances in Neural Information Processing Systems
  (NeurIPS)}, 2019.

\bibitem[Khirirat et~al.(2021)Khirirat, Magn{\'u}sson, Aytekin, and
  Johansson]{khirirat2021flexible}
Sarit Khirirat, Sindri Magn{\'u}sson, Arda Aytekin, and Mikael Johansson.
\newblock A flexible framework for communication-efficient machine learning.
\newblock In \emph{AAAI Conference on Artificial Intelligence}, 2021.

\bibitem[Korhonen \& Alistarh(2021)Korhonen and Alistarh]{korhonen2021towards}
Janne~H Korhonen and Dan Alistarh.
\newblock Towards tight communication lower bounds for distributed
  optimisation.
\newblock In \emph{Advances in Neural Information Processing Systems
  (NeurIPS)}, 2021.

\bibitem[Korpelevich(1976)]{Kor76}
Galina~Michailovna Korpelevich.
\newblock The extragradient method for finding saddle points and other
  problems.
\newblock \emph{{\`E}konom. i matematicheskie metody}, 12:\penalty0 747--756,
  1976.

\bibitem[Kovalev et~al.(2022)Kovalev, Beznosikov, Sadiev, Persiianov,
  Richt{\'a}rik, and Gasnikov]{kovalev2022optimal}
Dmitry Kovalev, Aleksandr Beznosikov, Abdurakhmon Sadiev, Michael Persiianov,
  Peter Richt{\'a}rik, and Alexander Gasnikov.
\newblock Optimal algorithms for decentralized stochastic variational
  inequalities.
\newblock \emph{arXiv preprint arXiv:2202.02771}, 2022.

\bibitem[Krizhevsky(2009)]{CIFAR10}
Alex Krizhevsky.
\newblock Learning multiple layers of features from tiny images.
\newblock 2009.
\newblock Technical report, University of Toronto.

\bibitem[Levy et~al.(2018)Levy, Yurtsever, and Cevher]{LYC18}
Kfir~Yehuda Levy, Alp Yurtsever, and Volkan Cevher.
\newblock Online adaptive methods, universality and acceleration.
\newblock In \emph{Advances in Neural Information Processing Systems
  (NeurIPS)}, 2018.

\bibitem[Li et~al.(2014)Li, Andersen, Park, Smola, Ahmed, Josifovski, Long,
  Shekita, and Su]{Li14}
Mu~Li, David~G. Andersen, Jun~Woo Park, Alexander~J. Smola, Amr Ahmed, Vanja
  Josifovski, James Long, Eugene~J. Shekita, and Bor-Yiing Su.
\newblock Scaling distributed machine learning with the parameter server.
\newblock In \emph{USENIX Symposium on Operating Systems Design and
  Implementation (OSDI)}, 2014.

\bibitem[Loizou et~al.(2021)Loizou, Berard, Gidel, Mitliagkas, and
  Lacoste-Julien]{loizou2021stochastic}
Nicolas Loizou, Hugo Berard, Gauthier Gidel, Ioannis Mitliagkas, and Simon
  Lacoste-Julien.
\newblock Stochastic gradient descent-ascent and consensus optimization for
  smooth games: Convergence analysis under expected co-coercivity.
\newblock In \emph{Advances in Neural Information Processing Systems
  (NeurIPS)}, 2021.

\bibitem[Malitsky(2020)]{Mal20}
Yura Malitsky.
\newblock Golden ratio algorithms for variational inequalities.
\newblock \emph{Mathematical Programming}, 184:\penalty0 383--410, 2020.

\bibitem[Markov et~al.(2022)Markov, Ramezanikebrya, and Alistarh]{CGX}
Ilia Markov, Hamidreza Ramezanikebrya, and Dan Alistarh.
\newblock Project {CGX}: Algorithmic and system support for scalable deep
  learning on a budget.
\newblock \emph{arXiv.2111.08617}, 2022.

\bibitem[Nemirovski(2004)]{Nem04}
Arkadi~Semen Nemirovski.
\newblock Prox-method with rate of convergence ${O}(1/t)$ for variational
  inequalities with {Lipschitz} continuous monotone operators and smooth
  convex-concave saddle point problems.
\newblock \emph{SIAM Journal on Optimization}, 15\penalty0 (1):\penalty0
  229--251, 2004.

\bibitem[Nemirovski et~al.(2009)Nemirovski, Juditsky, Lan, and Shapiro]{NJLS09}
Arkadi~Semen Nemirovski, Anatoli Juditsky, Guanghui Lan, and Alexander Shapiro.
\newblock Robust stochastic approximation approach to stochastic programming.
\newblock \emph{SIAM Journal on Optimization}, 19\penalty0 (4):\penalty0
  1574--1609, 2009.

\bibitem[Nesterov(2004)]{Nes04}
Yurii Nesterov.
\newblock \emph{Introductory Lectures on Convex Optimization: A Basic Course}.
\newblock Kluwer Academic Publishers, 2004.

\bibitem[Nesterov(2007)]{Nes07}
Yurii Nesterov.
\newblock Dual extrapolation and its applications to solving variational
  inequalities and related problems.
\newblock \emph{Mathematical Programming}, 109\penalty0 (2):\penalty0 319--344,
  2007.

\bibitem[Nesterov(2009)]{Nes09}
Yurii Nesterov.
\newblock Primal-dual subgradient methods for convex problems.
\newblock \emph{Mathematical Programming}, 120\penalty0 (1):\penalty0 221--259,
  2009.

\bibitem[Pinto et~al.(2017)Pinto, Davidson, Sukthankar, and
  Gupta]{pinto2017robust}
Lerrel Pinto, James Davidson, Rahul Sukthankar, and Abhinav Gupta.
\newblock Robust adversarial reinforcement learning.
\newblock In \emph{International Conference on Machine Learning (ICML)}, 2017.

\bibitem[Polyak(1987)]{Pol87}
Boris~Teodorovich Polyak.
\newblock \emph{Introduction to Optimization}.
\newblock Optimization Software, 1987.

\bibitem[Popov(1980)]{Pop80}
Leonid~Denisovich Popov.
\newblock A modification of the {Arrow}\textendash{Hurwicz} method for search
  of saddle points.
\newblock \emph{Mathematical Notes of the Academy of Sciences of the USSR},
  28\penalty0 (5):\penalty0 845--848, 1980.

\bibitem[Rakhlin \& Sridharan(2013)Rakhlin and Sridharan]{RS13-NIPS}
Alexander Rakhlin and Karthik Sridharan.
\newblock Optimization, learning, and games with predictable sequences.
\newblock In \emph{Advances in Neural Information Processing Systems
  (NeurIPS)}, 2013.

\bibitem[Ramezani-Kebrya et~al.(2021)Ramezani-Kebrya, Faghri, Markov, Aksenov,
  Alistarh, and Roy]{NUQSGD}
Ali Ramezani-Kebrya, Fartash Faghri, Ilya Markov, Vitalii Aksenov, Dan
  Alistarh, and Daniel~M. Roy.
\newblock {NUQSGD}: Provably communication-efficient data-parallel {SGD} via
  nonuniform quantization.
\newblock \emph{Journal of Machine Learning Research (JMLR)}, 22\penalty0
  (114):\penalty0 1--43, 2021.

\bibitem[Recht et~al.(2011)Recht, R\'{e}, Wright, and Niu]{Recht11}
Benjamin Recht, Christopher R\'{e}, Stephen~J. Wright, and Feng Niu.
\newblock {HOGWILD!}: A lock-free approach to parallelizing stochastic gradient
  descent.
\newblock In \emph{Advances in Neural Information Processing Systems
  (NeurIPS)}, 2011.

\bibitem[Sa et~al.(2015)Sa, Zhang, Olukotun, and R\'{e}]{Sa}
Christopher M.~De Sa, Ce~Zhang, Kunle Olukotun, and Christopher R\'{e}.
\newblock Taming the wild: A unified analysis of {HOGWILD!}-style algorithms.
\newblock In \emph{Advances in Neural Information Processing Systems
  (NeurIPS)}, 2015.

\bibitem[Schmidt et~al.(2018)Schmidt, Santurkar, Tsipras, Talwar, and
  Madry]{schmidt2018adversarially}
Ludwig Schmidt, Shibani Santurkar, Dimitris Tsipras, Kunal Talwar, and
  Aleksander Madry.
\newblock Adversarially robust generalization requires more data.
\newblock In \emph{Advances in Neural Information Processing Systems
  (NeurIPS)}, 2018.

\bibitem[Seide et~al.(2014)Seide, Fu, Droppo, Li, and Yu]{Seide14}
Frank Seide, Hao Fu, Jasha Droppo, Gang Li, and Dong Yu.
\newblock 1-bit stochastic gradient descent and its application to
  data-parallel distributed training of speech {DNN}s.
\newblock In \emph{INTERSPEECH}, 2014.

\bibitem[Shalev-Shwartz \& Ben-David(2014)Shalev-Shwartz and
  Ben-David]{shalev2014understanding}
Shai Shalev-Shwartz and Shai Ben-David.
\newblock \emph{Understanding machine learning: From theory to algorithms}.
\newblock Cambridge university press, 2014.

\bibitem[Stonyakin et~al.(2018)Stonyakin, Gasnikov, Dvurechensky, Alkousa, and
  Titov]{SGDA+18}
Fedor Stonyakin, Alexander Gasnikov, Pavel Dvurechensky, Mohammad Alkousa, and
  Alexander Titov.
\newblock Generalized mirror prox for monotone variational inequalities:
  {Universality} and inexact oracle, 2018.

\bibitem[Strom(2015)]{Strom15}
Nikko Strom.
\newblock Scalable distributed {DNN} training using commodity {GPU} cloud
  computing.
\newblock In \emph{INTERSPEECH}, 2015.

\bibitem[Syrgkanis et~al.(2015)Syrgkanis, Agarwal, Luo, and
  Schapire]{syrgkanis2015fast}
Vasilis Syrgkanis, Alekh Agarwal, Haipeng Luo, and Robert~E. Schapire.
\newblock Fast convergence of regularized learning in games.
\newblock In \emph{Advances in Neural Information Processing Systems
  (NeurIPS)}, 2015.

\bibitem[Tsitsiklis \& Luo(1987)Tsitsiklis and
  Luo]{tsitsiklis1987communication}
John~N. Tsitsiklis and Zhi-Quan Luo.
\newblock Communication complexity of convex optimization.
\newblock \emph{Journal of Complexity}, 3\penalty0 (3):\penalty0 231--243,
  1987.

\bibitem[Wang \& Joshi(2019)Wang and Joshi]{wang2019adaptive}
Jianyu Wang and Gauri Joshi.
\newblock Adaptive communication strategies to achieve the best error-runtime
  trade-off in local-update {SGD}.
\newblock In \emph{Proceedings of Machine Learning and Systems (MLSys)}, 2019.

\bibitem[Wen et~al.(2017)Wen, Xu, Yan, Wu, Wang, Chen, and Li]{TernGrad}
Wei Wen, Cong Xu, Feng Yan, Chunpeng Wu, Yandan Wang, Yiran Chen, and Hai Li.
\newblock {TernGrad}: Ternary gradients to reduce communication in distributed
  deep learning.
\newblock In \emph{Advances in Neural Information Processing Systems
  (NeurIPS)}, 2017.

\bibitem[Wilson et~al.(2017)Wilson, Roelofs, Stern, Srebro, and
  Recht]{wilson2017marginal}
Ashia~C. Wilson, Rebecca Roelofs, Mitchell Stern, Nati Srebro, and Benjamin
  Recht.
\newblock The marginal value of adaptive gradient methods in machine learning.
\newblock In \emph{Advances in Neural Information Processing Systems
  (NeurIPS)}, 2017.

\bibitem[Woodworth et~al.(2021)Woodworth, Bullins, Shamir, and
  Srebro]{woodworth2021min}
Blake~E Woodworth, Brian Bullins, Ohad Shamir, and Nathan Srebro.
\newblock The min-max complexity of distributed stochastic convex optimization
  with intermittent communication.
\newblock In \emph{Conference on Learning Theory (COLT)}, 2021.

\bibitem[Xing et~al.(2015)Xing, Ho, Dai, Kim, Wei, Lee, Zheng, Xie, Kumar, and
  Yu]{Petuum}
Eric~P. Xing, Qirong Ho, Wei Dai, Jin~Kyu Kim, Jinliang Wei, Seunghak Lee, Xun
  Zheng, Pengtao Xie, Abhimanu Kumar, and Yaoliang Yu.
\newblock Petuum: A new platform for distributed machine learning on big data.
\newblock \emph{{IEEE} transactions on Big Data}, 1(2):\penalty0 49--67, 2015.

\bibitem[Zhang et~al.(2017)Zhang, Li, Kara, Alistarh, Liu, and Zhang]{ZipML}
Hantian Zhang, Jerry Li, Kaan Kara, Dan Alistarh, Ji~Liu, and Ce~Zhang.
\newblock {ZipML}: Training linear models with end-to-end low precision, and a
  little bit of deep learning.
\newblock In \emph{International Conference on Machine Learning (ICML)}, 2017.

\bibitem[Zhang et~al.(2015)Zhang, Choromanska, and LeCun]{Zhang15}
Sixin Zhang, Anna Choromanska, and Yann LeCun.
\newblock Deep learning with elastic averaging {SGD}.
\newblock In \emph{Advances in Neural Information Processing Systems
  (NeurIPS)}, 2015.

\bibitem[Zhou et~al.(2018)Zhou, Wu, Ni, Zhou, Wen, and Zou]{Dorefa}
Shuchang Zhou, Yuxin Wu, Zekun Ni, Xinyu Zhou, He~Wen, and Yuheng Zou.
\newblock Dorefa-net: Training low bitwidth convolutional neural networks with
  low bitwidth gradients.
\newblock \emph{arXiv:1606.06160}, 2018.

\bibitem[Zinkevich et~al.(2010)Zinkevich, Weimer, Li, and Smola]{Zinkevich}
Martin~A. Zinkevich, Markus Weimer, Lihong Li, and Alex Smola.
\newblock Parallelized stochastic gradient descent.
\newblock In \emph{Advances in Neural Information Processing Systems
  (NeurIPS)}, 2010.

\end{thebibliography}

\appendix
\section{Appendix}\label{app:init}
\paragraph{Notation.} We use $\E[\cdot]$, $\|\cdot\|$, $\|\cdot\|_0$, and $\|\cdot\|_{\ast}$  to denote the expectation operator, Euclidean norm, number of nonzero elements of a vector, and dual norm, respectively. We use $|\cdot|$ to denote the length of a binary string, the length of a vector, and cardinality of a set. We use lower-case bold letters to denote vectors. Sets are typeset in a calligraphic font. The base-$2$ logarithm is denoted by $\log$, and the set of binary strings is denoted by $\{0,1\}^*$. We use $[n]$ to denote $\{1,\ldots,n\}$ for an integer $n$.

\paragraph{Content of the appendix.} The appendix is organized as follows: 
\begin{itemize}
    \item Complete related work is discussed  in~\cref{app:relatedwork}.
    \item Special cases of \ref{eq:QGEG} are provided in~\cref{app:special}.
    \item \cref{thm:varbound} (variance bound) is proved in~\cref{app:pr_var}.
    \item \cref{thm:codebound} (code-length bound) is proved in~\cref{app:pr_code}.
    \item \cref{thm:convQabs} (\ref{eq:QGEG} under absolute noise) is proved in~\cref{app:convQabs}.
    \item \cref{thm:convQrel} (\ref{eq:QGEG} under relative noise) is proved in~\cref{app:convQrel}.
    \item Additional experimental details are included in~\cref{app:experiment}. 
    \item Trade-off between number of iterations and time per iteration is provided in ~\cref{app:tradeoff}.
    \item Popular Examples motivating \cref{assumption:relative} are provided in~\cref{app:example}.
    \item The encoding/decoding details are provided in~\cref{app:encoding}.
\end{itemize}

\section{Further Related work}\label{app:relatedwork}
\paragraph{Unbiased  compression.} \citet{Seide14} proposed SignSGD, an efficient heuristic scheme to reduce communication costs drastically by quantizing each gradient component to two values. 
(This scheme is sometimes termed as  1bitSGD~\citep{Seide14}.)
\citet{signSGD} later provided convergence guarantees for a variant of SignSGD. 
Note that the quantization employed by SignSGD is not unbiased, and so a new analysis was required. \citet{QSGD} proposed quantized \sgd (QSGD) focusing on the uniform quantization of stochastic 
gradients normalized to have unit Euclidean norm. Their experiments illustrate a similar quantization method, where gradients are normalized to have unit $L^\infty$ norm, achieves better performance. We refer to this method as \qinf or Qinf in short. \citet{TernGrad} proposed TernGrad, which can be viewed as a special case of \qinf with three quantization levels. \citet{NUQSGD} proposed nonuniform quantization levels (\nuq) and demonstrated superior empirical results compared to \qinf. Recently, lattice-based quantization has been studied for distributed mean estimation and variance reduction \citep{lattice}.

Adaptive quantization has been used for speech communication and storage \citep{Cummiskey}. In machine learning, several biased and unbiased  schemes have been proposed to compress networks and gradients. In this work, we focus on unbiased and coordinate-wise schemes to compress gradients. \citet{ZipML} proposed ZipML, which is an optimal quantization method if all points to be quantized
are known a priori. To find the optimal sequence of 
quantization levels, a dynamic program is solved whose computational and 
memory cost is quadratic in the number of points to be quantized, which in the case of gradients would correspond to their dimension. %
\citet{ALQ} have proposed two adaptive gradient compression schemes where multiple processors update their compression schemes in parallel by efficiently computing sufficient statistics of a parametric
distribution. Adaptive quantization methods in~\citep{ALQ} are applicable only when  minimizing a single empirical risk. In addition, convergence guarantees in~\citep{ALQ} are established for smooth nonconvex optimization with a fixed step-size. In this paper, we propose  communication-efficient variants of generalized EG family of algorithms with an adaptive and unbiased quantization scheme for a general \eqref{eq:VI} problem. Furthermore, we establish improved variance and code-length bounds and optimal convergence guarantees for nonconvex problems with an adaptive step-size.

Adaptive gradient compression has been studied in other contexts when minimizing a single empirical risk, such as adapting the communication frequency in  local SGD~\citep{wang2019adaptive}, adapting the number of quantization levels (communication budget) over the course of training~\citep{guo2020accelerating,agarwal2021adaptive},~adapting a gradient sparsification scheme over the course of training~\citep{khirirat2021flexible}, and adapting compression parameters across
model layers and training iterations~\citep{CGX}. We focus on unbiased and normalized quantization and adapt quantization levels to {\it minimize quantization error} for generalized EG family of algorithms, which has not been considered in the literature. 

\paragraph{First-order methods to solve~\ref{eq:VI}s.}
In the \ref{eq:VI} ~literature, the benchmark method is  extra-gradient~(\EG), proposed by \citet{Kor76}, along with its variants including~ \citep{Nem04,Nes07}. Furthermore, there have been a line of work that aims to focus on establishing convergence guarantees though adopting an adaptive step-size policy. To that end, we review the most relevant works below.

For unconstrained problems with an operator that is locally Lipschitz continuous (but not necessarily globally), the Golden Ratio Algorithm (GRAAL) of~\citet{Mal20} achieves convergence without requiring prior knowledge of the problem's Lipschitz parameter.
Moreover, such guarantees are provided in problems with a bounded domain by the Generalized Mirror Prox (GMP) algorithm of \citet{SGDA+18} under the umbrella of H\"{o}lder continuity.

A more relevant method that simultaneously achieves an $\mathcal{O}(1/\sqrt{T})$ rate in non-smooth and/or stochastic  problems and an $\mathcal{O}(1/T)$ rate in smooth ones is the recent algorithm of \citet{BL19}.
This algorithm employs an adaptive, AdaGrad-like step-size policy which allows the method to interpolate between these  regimes.
On the negative side, this algorithm  requires a bounded domain with a (Bregman) diameter that is known in advance.

In optimization community, similar rate interpolation guarantees between different noise profiles has been explored by \citet{Universal}.
However, their results are limited to {\it centralized, single GPU settings}.
 
 While all these algorithms concern standard single-GPU settings with having access to full-precision stochastic dual vectors, we consider multi-GPU settings, which has not been considered before. 

 \citet{beznosikov2021distributed,kovalev2022optimal} have proposed communication-efficient algorithms for \ref{eq:VI}s with finite-sum structure and variance reduction in centralized settings and (strongly) monotone \ref{eq:VI}s in decentralized settings, respectively. Unlike \citep{beznosikov2021distributed,kovalev2022optimal}, we achieve fast and order-optimal rates with {\it adaptive step-size} and {\it adaptive compression}  without requiring variance reduction and strong monotoncity, and improve variance and code-length bounds for unbiased and adaptive compression.

\paragraph{Detailed comparison with~\citep{Universal,QSGD,ALQ,NUQSGD}.}%
In this section, we elaborate and provide a detailed comparison with the most relevant related work~\citep{Universal,QSGD,ALQ,NUQSGD}. 

Although our results build upon the extra-gradient literature~\citet{Kor76}, as~\citet{Universal} does, in this paper, we address {\it monotone VI/ convex-concave min-max problems} in {\it distributed and large-scale settings}, which has not been considered in \citet{Universal} that strictly refers to a strictly single-GPU and centralized setting. The considered distributed framework complicates the analysis in a significant manner, since we have to simultaneously treat two different {\it types} of randomness. In particular, on one hand, we face randomness associated with the compression scheme (which is necessary to achieve substantial communication savings in a distributed setup) where on the other hand we deal with different noisy feedback models stemming from inexact operator calculations (before any compression takes place), which together result in efficient implementations at each GPU. We show benefits of distributed training in terms of accelerating convergence for general monotone \ref{eq:VI}s. 

Unlike~\citep[Theorem 3.2]{QSGD} and~\citep[Theorem 4]{NUQSGD} that hold under the special cases of $L^2$ normalization with uniform and exponentially spaced levels, respectively, our bound in~\cref{thm:varbound} holds under general $L^q$ normalization and arbitrary sequence of quantization levels. For the special case of $L^2$ normalization in the regime of large $d$, which is the case in practice, our bound in~\cref{thm:varbound} is ${\Oc} (\ell_1\sqrt{d})$, which is arbitrarily smaller than ${\Oc} (\sqrt{d}/s)$ and ${\Oc} (2^{-s}\sqrt{d})$ in \citep[Theorem 3.2]{QSGD} and \citep[Theorem 4]{NUQSGD}, respectively, because $\ell_1$ is adaptively designed to minimize the variance of quantization. Furthermore, unlike the bound in \citep[Theorem 2]{ALQ} that requires an inner optimization problem over an auxiliary parameter $p$, the bound in~\cref{thm:varbound} is provided in an explicit form without an inner optimization problem, which matches the lower bound. 

Unlike~\citep[Theorem 3.4]{QSGD} and ~\citep[Theorem 5]{NUQSGD} that hold under the special cases of $L^2$ normalization, our code-length bound in~\cref{thm:codebound} holds under general $L^q$ normalization.  For the special case of $L^2$ normalization with $s=\sqrt{d}$ as in~\citep[Theorem 3.4]{QSGD}, our bound in~\cref{thm:codebound} can be arbitrarily smaller than~\citep[Theorem 3.4]{QSGD} and ~\citep[Theorem 5]{NUQSGD} depending on $\{p_0,\ldots,p_{s+1}\}$. Compared to \citep[Theorem 3]{ALQ}, our bound in~\cref{thm:codebound} does not have an additional $n_{\ell_1,d}$ term.

\section{Special cases of \ref{eq:QGEG}}\label{app:special}
In this section, we show that under different choices of $V_{k,t}$ and $V_{k,t+1/2}$, one can obtain {\it communication-efficient} variants of stochastic dual averaging~\citep{Nes09}, stochastic dual extrapolation~\citep{Nes07}, and stochastic optimistic dual averaging~\citep{Pop80,RS13-NIPS,HAM21,HIMM22} in {\it multi-GPU settings} as special cases of \ref{eq:QGEG}.

\begin{example}{\bf{Communication-efficient stochastic dual averaging:}}
Consider the case where $\hat V_{k,t}\equiv0$ and $\hat V_{k,t+1/2}\equiv \hat g_{k,t+1/2}=Q_{\bql_t}(A(X_{t+1/2})+U_{k,t+1/2})$. This setting yields to $X_{t+1/2}=X_{t}$ and hence $\hat g_{k,t+1/2}=\hat g_{k,t}=\hat V_{k,t+1/2}$. Therefore, \ref{eq:QGEG} reduces to the communication-efficient stochastic dual averaging scheme:
\begin{align}
    \label{eq:DA}
    \tag{Quantized DA}
    \begin{split}
    Y_{t+1}&=Y_{t}-K^{-1}\sum_{k=1}^K\hat g_{k,t}
    \\
    X_{t+1}&=\gamma_{t+1}Y_{t+1}
    \end{split}
\end{align}  
\end{example}  
\begin{example}{\bf{Communication-efficient stochastic dual extrapolation}:}
Consider the case where $\hat V_{k,t}\equiv \hat g_{k,t}=Q_{\bql_t}(A(X_t)+U_{k,t})$ and $\hat V_{k,t+1/2}\equiv \hat g_{k,t+1/2}=Q_{\bql_t}(A(X_{t+1/2})+U_{k,t+1/2})$ are noisy oracle queries at $X_{t}$ and $X_{t+1/2}$, respectively.
Then \ref{eq:QGEG} provides the communication-efficient variant of Nesterov's stochastic dual extrapolation method \citep{Nes07}:
\begin{align}
    \label{eq:DE}
    \tag{Quantized DE}
    \begin{split}
    X_{t+1 / 2}&=X_{t}-\frac{\g_t}{K}\sum_{k=1}^K  \hat g_{k,t}\\
    Y_{t+1}&=Y_{t}-K^{-1}\sum_{k=1}^K\hat g_{k,t+1 / 2}\\
    X_{t+1}&=\gamma_{t+1} Y_{t+1}
    \end{split}
\end{align}
\end{example}
\begin{example}{\bf{Communication-efficient stochastic optimistic dual averaging}:}
Consider the case $\hat V_{k,t}\equiv \hat g_{k,t-1/2}=Q_{\bql_t}(A(X_{t-1/2})+U_{k,t-1/2})$ and $\hat V_{k,t+1/2}\equiv\hat g_{k,t+1/2}=Q_{\bql_t}(A(X_{t+1/2})+U_{k,t+1/2})$ are the noisy oracle feedback at $X_{t-1/2}$ and $X_{t+1/2}$, respectively.
We then obtain the communication-efficient stochastic optimistic dual averaging method:
\begin{align}
    \label{eq:OptDA}
    \tag{Quantized OptDA}
    \begin{split}
    X_{t+1 / 2}&=X_{t}-\frac{\g_t}{K}\sum_{k=1}^K  \hat g_{k,t-1/2} \\
    Y_{t+1}&=Y_{t}-K^{-1}\sum_{k=1}^K\hat g_{k,t+1 / 2} \\
    X_{t+1}&=\gamma_{t+1} Y_{t+1}
\end{split}
\end{align}
\end{example}

\section{Proof of \cref{thm:varbound} (variance bound)}\label{app:pr_var}
Let $u_j=|v_j|/\|\vbf\|_q$, $\Bc_0 :=[0,\ql_1]$, and $\Bc_{j}:=[\ql_j,\ql_{j+1}]$ for $j\in[s]$. Let ${\tt V}_{\bql}(\vbf)=\E_{\qbf_\bql}[\|Q_{\bql}(\vbf)-\vbf\|_2^2]$ denote the variance of quantization in~\cref{var}. Then we have 

\begin{align}\label{var_reform}
{\tt V}_{\bql}(\vbf)=\|\vbf\|_q^2\big(\sum_{u_i\in\Bc_0}(\ql_1-u_i)u_i+\sum_{j=1}^{s}\sum_{u_i\in\Bc_{j}}(\ql_{j+1}-u_i)(u_i-\ql_j)\big).
\end{align} 

We first find the minimum $k_j$ that satisfies $(\ql_{j+1}-u)(u-\ql_{j})\leq k_ju^2$ for $u\in \Bc_{j}$ and $j\in[s]$. The minimum $k_j$ can be obtained by changing the variable $u=\ql_j\theta$:

\begin{align}
\begin{split}\label{k_for_r2}
k_j &= \max_{1\leq\theta\leq \ql_{j+1}/\ql_j}\frac{(\ql_{j+1}/\ql_j-\theta)(\theta-1)}{\theta^2}\\
&=\frac{\big(\ql_{j+1}/\ql_{j}-1\big)^2}{4(\ql_{j+1}/\ql_{j})}.
\end{split} 
\end{align}

We note that $\ql_{j+1}/\ql_j>1$ and $(x-1)^2/(4x)$ is monotonically increasing function of $x$ for $x>1$.

Furthermore, note that 
\begin{align}\nn
\sum_{u_i\notin\Bc_{0}}u_i^2\leq \frac{\|\vbf\|_2^2}{\|\vbf\|_q^2}. \end{align}

Substituting \cref{k_for_r2} into \cref{var_reform}, an upper bound on ${\tt V}_{\bql}(\vbf)$ is given by 
\begin{align}\nn
{\tt V}_{\bql}(\vbf)\leq \|\vbf\|_q^2\Big(\Big(\frac{\ol\ql+\ol\ql^{-1}}{4}-\frac{1}{2}\Big)\frac{\|\vbf\|_2^2}{\|\vbf\|_q^2}+\sum_{u_i\in\Bc_0}(\ql_1-u_i)u_i\Big).
\end{align}

In the rest of the proof, we use the following known lemma. 
\blm\label{lm:normineq} 
Let $\vbf\in\reals^d$. Then, for all $0<p<q$, we have $\|\vbf\|_q\leq \|\vbf\|_p\leq d^{1/p-1/q}\|\vbf\|_q$.  
\elm 
We note that  \cref{lm:normineq} holds even when $q<1$ and $\|\cdot\|_q$ is merely a seminorm.

We now establish an upper bound on $\sum_{u_i\in\Bc_0}(2^{-s}-u_i)u_i$.

\blm[{\citealt[Lemma 15]{NUQSGD}}]\label{lm:K_p}
Let $p\in (0,1)$ and $u\in \Bc_0$. Then we have $u(\ql_1-u)\leq K_p{\ql_1}^{(2-p)} u^p$ where 
\begin{align}\label{K_p}
K_p=\Big(\frac{1/p}{2/p-1}\Big)\Big(\frac{1/p-1}{2/p-1}\Big)^{(1-p)}. 
\end{align}
\elm

Let ${\Sc_j}$ denote the coordinates of vector ${\vbf}$ whose elements fall into the ${(j+1)}$-th bin, \ie ${\Sc_j:=\{i:u_i\in\Bc_j\}}$ for  $j\in[s]$. %
For any $0<p<1$ and $q\geq 2$, we have  
\begin{align}
\|\vbf\|_q^{2}\sum_{u_i\in\Bc_0}u_i^p&= \|\vbf\|_q^{2-p}\sum_{i\in\Sc_0}|v_i|^p\nn\\
&\leq \|\vbf\|_q^{2-p}\|\vbf\|_p^p\nn\\
&\leq \|\vbf\|_q^{2-p}\|\vbf\|_2^pd^{1-p/2}\nn\\
&\leq \|\vbf\|_2^2d^{1-p/2}\nn,
\end{align} where the third inequality holds as $\|\vbf\|_p\leq \|\vbf\|_2d^{1/p-1/2}$ using \cref{lm:normineq} and the last inequality holds as $\|\vbf\|_q\leq \|\vbf\|_2$ for $q\geq 2$. Using \cref{lm:normineq,lm:K_p}, we establish an upper bound on ${\tt V}_{\bql}(\vbf)$:  
\begin{align}\nn
{\tt V}_{\bql}(\vbf)\leq \|\vbf\|_2^2\Big(\frac{\ol\ql+\ol\ql^{-1}}{4}-\frac{1}{2}+K_p{\ql_1}^{(2-p)}d^{1-p/2}\Big). 
\end{align}

For $q\geq 1$, we note that $\|\vbf\|_q^{2-p}\leq \|\vbf\|_2^{2-p}d^{\frac{2-p}{\min\{q,2\}}-\frac{2-p}{2}}$, \ie
\begin{align}\label{var:Kp}
{\tt V}_{\bql}(\vbf)\leq \|\vbf\|_2^2\Big(\frac{\ol\ql+\ol\ql^{-1}}{4}-\frac{1}{2}+K_p{\ql_1}^{(2-p)}d^{\frac{2-p}{\min\{q,2\}}}\Big). 
\end{align}

Note that the optimal $p$ to minimize $\epsilon_Q$ is obtained by minimizing:  
\begin{align}\nn
\lambda(p) =  \big(\frac{1/p}{2/p-1}\big)\big(\frac{1/p-1}{2/p-1}\big)^{1-p}\delta^{1-p}
\end{align} where $\delta = \ql_1 d^{\frac{1}{\min\{q,2\}}}$. 

Taking the first-order derivative of $\lambda(p)$, the optimal $p^*$ is given by  
\begin{align}\label{optpcaes}
p^*=
\begin{cases}
\frac{\delta-2}{\delta-1},\quad\delta\geq 2 \\
0,\quad \delta< 2. 
\end{cases}
\end{align}
Substituting~\eqref{optpcaes} into~\eqref{var:Kp} gives \eqref{varbound}, which completes the proof. 

\section{Proof of \cref{thm:codebound} (code-length bound)}\label{app:pr_code}

Let $|\cdot|$ denote the length of a binary string.
In this section, we obtain an upper bound on $\E_{\o}\E_{\qbf_\bql}[|\ENCODE(Q_\bql(g(\xbf;\o));\bql)|]$, \ie the expected number of communication bits per iteration. 
We recall from \cref{sec:encoding} that  $\vbf$ is uniquely represnted by the tuple 
$(\|\vbf\|_q,\signvec,\ubf)$. We first encode the norm $\|\vbf\|_q$ using $C_b$ bits where. In practice, we use standard 32-bit floating point encoding.

We then use one bit to encode the sign of each nonzero entry of $\ubf$. Let ${\Sc_j:=\{i:u_i\in[l_j,l_{j+1}]\}}$ and $N_j:= |\Sc_j|$ for  ${j\in[s]}$. We now provide the expression for  probabilities associated with our symbols to be coded, \ie $\{\ql_0,\ql_1,\ldots,\ql_{s+1}\}$. The associated probabilities can computed using the weighted sum of the conditional CDFs of normalized coordinates in~\ref{Qadap} and quantization levels: 

\bprop\label{prop:codePMF} Let $j\in[s]$. The probability of occurrence of $\ql_j$ (weight of symbol $\ql_j$) is given by 
\begin{align}\nn%
p_j=\Pr(\ql_j)=\int_{\ql_j-1}^{\ql_j}\frac{u-\ql_{j-1}}{\ql_{j}-\ql_{j-1}}\ud\tilde\cdf(u)+\int_{\ql_j}^{\ql_{j+1}}\frac{\ql_{j+1}-u}{\ql_{j+1}-\ql_{j}}\ud\tilde\cdf(u)
\end{align}  where $\tilde F$ is the weighted sum of the conditional CDFs of normalized coordinates in~\ref{Qadap}. In addition, we have 
\begin{align}
p_0=\Pr(\ql_0=0)=\int_{0}^{\ql_1}\frac{1-u}{\ql_1}\ud\tilde\cdf(u)~\and~p_{s+1}=\Pr(\ql_{s+1}=1)=\int_{\ql_s}^{1}\frac{u-\ql_{s}}{1-\ql_{s}}\ud\tilde\cdf(u).\nn
\end{align}
\eprop

We have an upper bound on the expected number of nonzero entries as follows:

\blm\label{lm:sparsity}
Let $\vbf\in\reals^d$. The expected number of nonzeros in $Q_{\bql}(\vbf)$ is given by
\begin{align}
    \E_{\qbf_\bql}[\|Q_{\bql}(\vbf)\|_0]=(1-p_0)d.
\end{align}
\elm 

We then send the the associated codeword to encode each coordinate of $\ubf$. The optimal expected code-length for transmitting one random symbol is within one bit of the entropy of the source~\citep{InfTheory}. So the number of required information bits to transmit entries of $\ubf$ is bounded above by $d(H(L)+1)$ where $H(L)=-\sum_{j=1}^sp_j\log(p_j)$ is the entropy in bits. Putting everything together, we have 
\begin{align}\nn
\E_{\o}\E_{\qbf_\bql}[|\ENCODE(Q_\bql(g(\xbf;\o));\bql)|]\leq C_b+(1-p_0)d+(H(L)+1)d
\end{align} where $C_b=\bigoh(1)$ is a universal constant.
Finally, we note that the entropy of a source with $n$ outcomes is upper bounded by $\log(n)$.

\section{Proof of \cref{thm:convQabs} (\ref{eq:QGEG} under absolute noise)}\label{app:convQabs}

We note that the output of~\cref{QGEGalg} follows the iterates of~\eqref{eq:QGEG}:  
\begin{align}
    \tag{\QEG}
    \begin{split}
    X_{t+1 / 2}&=X_{t}-\frac{\gamma_t}{K}\sum_{k=1}^K  \hat V_{k,t} \\
    Y_{t+1}&=Y_{t}-K^{-1}\sum_{k=1}^K\hat V_{k,t+1 / 2} \\
    X_{t+1}&=\gamma_{t+1} Y_{t+1}
    \end{split}
\end{align}

We first prove the following~{\it Template Inequality} for~\eqref{eq:QGEG}, which is a useful milestone to prove both~\cref{thm:convQabs,thm:convQrel}. 

\bprop[Template inequality]\label{prop:TemIneq} Let $X\in\reals^d$. Suppose the iterates $X_t$ of~\eqref{eq:QGEG} are updated with some {\it non-increasing} step-size schedule $\g_t$ for $t=1,1/2,\ldots$ Then, we have 
\begin{align}\label{eq:TemIneq}
\begin{split}
\sum_{t=1}^T\Big\langle\frac{1}{K}\sum_{k=1}^K\hat V_{k,t+1 / 2},X_{t+1 / 2}-X \Big\rangle &\leq\frac{\|X\|_{\ast}^2}{2\g_{T+1}}+\frac{1}{ 2K^2}\sum_{t=1}^T\g_t\sum_{k=1}^K\|\hat V_{k,t+1 / 2}-\hat V_{k,t}\|_{\ast}^2\\&\quad -\frac{1}{2}\sum_{t=1}^T\frac{1}{\g_t}\|X_t-X_{t+1 / 2}\|_{\ast}^2.    
\end{split}
\end{align}
\eprop
\bpr

We first decompose $\frac{1}{K}\langle\sum_{k=1}^K\hat V_{k,t+1 / 2},X_{t+1 / 2}-X \rangle$ into two terms and note: \begin{align}\nn
\begin{split}
\frac{1}{K}\Big\langle\sum_{k=1}^K\hat V_{k,t+1 / 2},X_{t+1 / 2}-X \Big\rangle &= S_A+S_B.
\end{split}
\end{align}
where 
\begin{align}\nn
    S_A =  \frac{1}{K}\Big\langle\sum_{k=1}^K\hat V_{k,t+1 / 2},X_{t+1 / 2}-X_{t+1} \Big\rangle   
\end{align}
and 
\begin{align}\nn
    S_B = \frac{1}{K}\Big\langle\sum_{k=1}^K\hat V_{k,t+1 / 2},X_{t+1}-X \Big\rangle.
\end{align}
Note that the update rule in~\eqref{eq:QGEG} implies:
\begin{align}\nn
\begin{split}
S_B &= \langle Y_{t}-Y_{t+1},X_{t+1}-X\rangle\\&=\Big\langle Y_{t}-\frac{\g_{t+1}}{\g_t}Y_{t+1},X_{t+1}-X\Big\rangle+\Big\langle \frac{\g_{t+1}}{\g_t}Y_{t+1}-Y_{t+1},X_{t+1}-X\Big\rangle\\&=\frac{1}{\g_t}\langle\g_t Y_{t}-\g_{t+1}Y_{t+1},X_{t+1}-X\rangle+\Big(\frac{1}{\g_{t+1}}-\frac{1}{\g_t}\Big)\langle-\g_{t+1}Y_{t+1},X_{t+1}-X\rangle\\&=
\frac{1}{\g_t}\langle X_{t}-X_{t+1},X_{t+1}-X\rangle+\Big(\frac{1}{\g_{t+1}}-\frac{1}{\g_t}\Big)\langle-X_{t+1},X_{t+1}-X\rangle.
\end{split}
\end{align}

By algebraic manipulations, we can further show that  
\begin{align}\nn
\begin{split}
S_B &=
\frac{1}{\g_t}
\l(\frac{1}{2}\|X_{t}-X\|_{\ast}^2-\frac{1}{2}\|X_{t+1}-X\|_{\ast}^2-\frac{1}{2}\|X_{t+1}-X_{t}\|_{\ast}^2\r)\\&\quad +\Big(\frac{1}{\g_{t+1}}-\frac{1}{\g_t}\Big)\l(\frac{1}{2}\|X\|_{\ast}^2-\frac{1}{2}\|X_{t+1}\|_{\ast}^2-\frac{1}{2}\|X_{t+1}-X\|_{\ast}^2\r)\\&\leq\frac{1}{2\g_t}
\|X_{t}-X\|_{\ast}^2-\frac{1}{2\g_{t+1}}\|X_{t+1}-X\|_{\ast}^2-\frac{1}{2\g_{t}}\|X_{t}-X_{t+1}\|_{\ast}^2+\Big(\frac{1}{2\g_{t+1}}-\frac{1}{2\g_t}\Big)\|X\|_{\ast}^2\end{split}
\end{align}
where the last inequality holds by dropping $-\frac{1}{2}\|X_{t+1}\|_{\ast}^2$. 

Rearranging the terms in the above expression and substituting $S_B$, we have 
\begin{align}\label{eq:PreTel1}
\begin{split}
\frac{1}{2\g_{t+1}}\|X_{t+1}-X\|_{\ast}^2&\leq 
\frac{1}{2\g_t}
\|X_{t}-X\|_{\ast}^2-\frac{1}{2\g_{t}}\|X_{t}-X_{t+1}\|_{\ast}^2+\Big(\frac{1}{2\g_{t+1}}-\frac{1}{2\g_t}\Big)\|X\|_{\ast}^2\\&\quad -\frac{1}{K}\Big\langle\sum_{k=1}^K\hat V_{k,t+1 / 2},X_{t+1}-X \Big\rangle\\&=\frac{1}{2\g_t}
\|X_{t}-X\|_{\ast}^2-\frac{1}{2\g_{t}}\|X_{t}-X_{t+1}\|_{\ast}^2+\Big(\frac{1}{2\g_{t+1}}-\frac{1}{2\g_t}\Big)\|X\|_{\ast}^2\\&\quad +\frac{1}{K}\Big\langle\sum_{k=1}^K\hat V_{k,t+1/2},X_{t+1/2}-X_{t+1}\Big\rangle-\frac{1}{K}\Big\langle\sum_{k=1}^K\hat V_{k,t+1 / 2},X_{t+1/2}-X \Big\rangle. 
\end{split}
\end{align}

On the other hand, we have 

\begin{align}\label{eq:appInPro}
\begin{split}
\frac{\g_t}{K}\Big\langle\sum_{k=1}^K\hat V_{k,t},X_{t+1/2}-X \Big\rangle&=\langle X_t-X_{t+1/2},X_{t+1/2}-X\rangle \\&=\frac{1}{2}
\|X_{t}-X\|_{\ast}^2-\frac{1}{2}\|X_{t}-X_{t+1/2}\|_{\ast}^2-\frac{1}{2}\|X_{t+1/2}-X\|_{\ast}^2 
\end{split}
\end{align}

Substituting $X=X_{t+1}$ and dividing both sides of~\eqref{eq:appInPro} by $\g_t$, we have 

\begin{align}\label{eq:PreTel2}
\begin{split}
\frac{1}{K}\Big\langle\sum_{k=1}^K\hat V_{k,t},X_{t+1/2}-X_{t+1} \Big\rangle&=\frac{1}{2\g_t}
\|X_{t}-X_{t+1}\|_{\ast}^2-\frac{1}{2\g_t}\|X_{t}-X_{t+1/2}\|_{\ast}^2\\&\quad-\frac{1}{2\g_t}\|X_{t+1/2}-X_{t+1}\|_{\ast}^2. 
\end{split}
\end{align}

Combining~\eqref{eq:PreTel1} and~\eqref{eq:PreTel2}, we have  

\begin{align}\nn
\begin{split}
\frac{1}{K}\Big\langle\sum_{k=1}^K\hat V_{k,t+1/2},X_{t+1/2}-X\Big\rangle&\leq\frac{1}{2\g_t}
\|X_{t}-X\|_{\ast}^2-\frac{1}{2\g_{t+1}}\|X_{t+1}-X\|_{\ast}^2+\big(\frac{1}{2\g_{t+1}}-\frac{1}{2\g_t}\big)\|X\|_{\ast}^2\\&\quad+\frac{1}{K}\Big\langle\sum_{k=1}^K\hat V_{k,t+1/2}-\hat V_{k,t},X_{t+1/2}-X_{t+1}\Big\rangle\\&\quad-\frac{1}{2\g_t}\|X_t-X_{t+1/2}\|_{\ast}^2-\frac{1}{2\g_t}\|X_{t+1}-X_{t+1/2}\|_{\ast}^2. 
\end{split}
\end{align}

Summing the above for $t=1,\ldots,T$ and telescoping, we have 
\begin{align}\nn\footnotesize 
\begin{split}
\frac{1}{K}\sum_{t=1}^T\langle\sum_{k=1}^K\hat V_{k,t+1/2},X_{t+1/2}-X\rangle&\leq\frac{1}{2\g_1}
\|X_{1}-X\|_{\ast}^2-\frac{1}{2\g_{T+1}}\|X_{T+1}-X\|_{\ast}^2+(\frac{1}{2\g_{T+1}}-\frac{1}{2\g_1})\|X\|_{\ast}^2\\&\quad+\frac{1}{K}\sum_{t=1}^T\langle\sum_{k=1}^K\hat V_{k,t+1/2}-\hat V_{k,t},X_{t+1/2}-X_{t+1}\rangle\\&\quad-\sum_{t=1}^T\frac{1}{2\g_t}\|X_t-X_{t+1/2}\|_{\ast}^2-\sum_{t=1}^T\frac{1}{2\g_t}\|X_{t+1}-X_{t+1/2}\|_{\ast}^2. 
\end{split}
\end{align}
Substituting $X_1=0$, we have\footnote{The substitution $X_1=0$ is just for notation simplicity and can be relaxed at the expense of obtaining a slightly more complicated expression. } 
\begin{align}\label{eq:InnerX}\footnotesize 
\begin{split}
\frac{1}{K}\sum_{t=1}^T\langle\sum_{k=1}^K\hat V_{k,t+1/2},X_{t+1/2}-X\rangle&\leq\frac{1}{2\g_{T+1}}
\|X\|_{\ast}^2+\frac{1}{K}\sum_{t=1}^T\langle\sum_{k=1}^K\hat V_{k,t+1/2}-\hat V_{k,t},X_{t+1/2}-X_{t+1}\rangle\\&\quad-\sum_{t=1}^T\frac{1}{2\g_t}\|X_t-X_{t+1/2}\|_{\ast}^2-\sum_{t=1}^T\frac{1}{2\g_t}\|X_{t+1}-X_{t+1/2}\|_{\ast}^2. 
\end{split}
\end{align}

By applying Cauchy–Schwarz and triangle inequalities, we have 
\begin{align}\small\label{eq:CSTriangle}
\begin{split}
\frac{1}{K}\langle\sum_{k=1}^K\hat V_{k,t+1/2}-\hat V_{k,t},X_{t+1/2}-X_{t+1}\rangle&\leq \sum_{k=1}^K 
\|\hat V_{k,t+1/2}-\hat V_{k,t}\|_{\ast}\|\frac{1}{K}(X_{t+1/2}-X_{t+1})\|_{\ast}.
\end{split}
\end{align}

Furthermore, since $ab\leq\frac{\g_t}{2K^2}a^2+\frac{K^2}{2\g_t}b^2$, we have \begin{align}\label{eq:partialInnerfixed}
\begin{split}
\sum_{t=1}^T\frac{1}{K}\langle\sum_{k=1}^K\hat V_{k,t+1/2}-\hat V_{k,t},X_{t+1/2}-X_{t+1}\rangle&\leq    \sum_{t=1}^T\frac{\g_t}{2K^2}\sum_{k=1}^K\|(\hat V_{k,t+1/2}-\hat V_{k,t})\|_{\ast}^2\\&\quad+\sum_{t=1}^T\frac{1}{2\g_t}\|X_{t+1/2}-X_{t+1}\|_{\ast}^2.
\end{split}
\end{align}

Substituting~\eqref{eq:partialInnerfixed} into~\eqref{eq:InnerX} and applying the convexity of $\|\cdot\|_{\ast}^2$, we obtain~\eqref{eq:TemIneq}, which completes the proof.  
\epr

The following lemmas show how additional noise due to compression affects the upper bounds under {\it absolute noise} and {\it relative noise} models in \cref{assumption:random,assumption:relative}, respectively.
Let $\qbf_\bql\sim\mathbb{P}_Q$ represent $d$ variables sampled i.i.d. for random quantization in \cref{def:quantf}. We remind that $\qbf_\bql$ is {\it independent} of the random sample $\o\sim\mathbb{P}$ in \cref{eq:oracle}.
We consider a general unbiased and normalized compression scheme, which has a bounded variance as in~\cref{thm:varbound}
and a bound on the expected number of communication bits to encode $Q_\bql(\vbf)$, \ie the output of $\ENCODE$ introduced in~\cref{sec:encoding}, which  satisfies~\cref{thm:codebound}. Then we have: 

\blm[ Unbiased compression under absolute noise]
\label{lm:abs} Let $\xbf\in\reals^d$ and $\o\sim\mathbb{P}$. Suppose the oracle $g(\xbf;\o)$ satisfies~\cref{assumption:random}. Suppose $Q_\bql$ satisfies~\cref{thm:varbound,thm:codebound}. Then the compressed $Q_\bql(g(\xbf;\o))$ satisfies~\cref{assumption:random} with  
\begin{align}
\E\l[\norm{Q_\bql(g(\xbf;\o))-A(\xbf)}_{2}^{2}\r]\leq \epsilon_QM^2+\s^2.
\end{align}
Furthermore,  the number of bits to encode $Q_\bql(g(\xbf;\o))$ is bounded by the upper bound in~\eqref{codebound}. 
\elm
\bpr
The almost sure boudedness and unbiasedness are immediately followed by the construction of  the unbiased $Q_\bql$. In particular, we note that the maximum additional norm when compressing $Q_\bql(g(\xbf;\o))$ happens when all normalized coordinates of $g(\xbf;\o)$, which we call them $u_1,\ldots,u_d$ are mapped to the upper level $\ql_{\level{u}+1}$~in~\cref{def:quantf}. The additional norm multiplier is bounded by $\sum_{i=1}^d(\ql_{\level{u_i}+1}-u_i)^2\leq \sum_{i=1}^du_i^2=1$. Then we have $Q_\bql(g(\xbf;\o))\leq 2M$ a.s., so the additional upper bound is constant. The final property also holds as follows: 
\begin{align}\nn
\begin{split}
\E_{\o}\E_{\qbf_{\bql}}\l[\norm{Q_\bql(g(\xbf;\o))-A(\xbf)}_{2}^{2}\r]&=\E_{\o}\E_{\qbf_{\bql}}\l[\norm{Q_\bql(g(\xbf;\o))\pm g(\xbf;\o)-A(\xbf)}_{2}^{2}\r]\\&= \E_{\o}\E_{\qbf_{\bql}}\l[\norm{Q_\bql(g(\xbf;\o))- g(\xbf;\o)}_{2}^{2}\r]+\E_{\o}\l[\norm{ U(\xbf;\o)}_{2}^{2}\r]\\&\leq \epsilon_Q\E_{\o}\l[\norm{g(\xbf;\o)}_{2}^{2}\r]+\s^2\\&\leq \epsilon_QM^2+\s^2
\end{split}
\end{align}
where the second step holds due to unbiasedness of $\qbf_{\bql}$ and the last inequality holds since $\|g(\xbf;\o)\|_{\ast}\leq M$ a.s. 
\epr 

\blm[Unbiased compression under relative noise]
\label{lm:rel} Let $\xbf\in\reals^d$ and $\o\sim\mathbb{P}$. Suppose the oracle $g(\xbf;\o)$ satisfies~\cref{assumption:relative}. Suppose $Q_\bql$ satisfies~\cref{thm:varbound,thm:codebound}. Then the compressed $Q_\bql(g(\xbf;\o))$ satisfies~\cref{assumption:relative} with \begin{align}
\E\l[\norm{Q_\bql(g(\xbf;\o))-A(\xbf)}_{2}^{2}\r]\leq \big(\epsilon_Q(c+1)+c\big)\norm{A(\xbf)}_{2}^{2}.
\end{align}
Furthermore,  the number of bits to encode $Q_\bql(g(\xbf;\o))$ is bounded by the upper bound in~\eqref{codebound}. 
\elm
\bpr The almost sure boudedness and unbiasedness are immediately followed by the construction of  the unbiased $Q_\bql$. The final property also holds as follows: 
\begin{align}\nn
\begin{split}
\E_{\o}\E_{\qbf_{\bql}}\l[\norm{Q_\bql(g(\xbf;\o))-A(\xbf)}_{2}^{2}\r]&=\E_{\o}\E_{\qbf_{\bql}}\l[\norm{Q_\bql(g(\xbf;\o))-g(\xbf;\o)}_{2}^{2}\r]+\E_{\o}\l[\norm{U(\xbf;\o)}_{2}^{2}\r]\\&\leq \epsilon_Q\E_{\o}\l[\norm{g(\xbf;\o)}_{2}^{2}\r]+c\norm{A(\xbf)}_{2}^{2}\\&=
\epsilon_Q\E_{\o}\l[\norm{U(\xbf;\o)+A(\xbf)}_{2}^{2}\r]+c\norm{A(\xbf)}_{2}^{2}\\&=
\epsilon_Q\big(\E_{\o}\l[\norm{U(\xbf;\o)}_{2}^{2}\r]+\norm{A(\xbf)}_{2}^{2}\big)+c\norm{A(\xbf)}_{2}^{2}\\&\leq \big(\epsilon_Q(c+1)+c\big)\norm{A(\xbf)}_{2}^{2}.
\end{split}  
\end{align} where the first and fourth steps hold due to unbiasedness of $\qbf_{\bql}$ and the noise model, respectively.
\epr

We now prove our main theorme: 

\bth[\ref{eq:QGEG} under absolute noise]%
Let $\Cc\subset\reals^d$ denote a compact neighborhood of a solution for \eqref{eq:VI} and let %
$D^2:=\sup_{X\in\Cc} \|X-X_0\|^2$. Suppose that the oracle and the problem~\eqref{eq:VI} satisfy Assumptions~\ref{assumption:ex} and~\ref{assumption:random}, respectively,~\cref{QGEGalg}  
is executed for $T$ iterations on $K$ processors with an adaptive step-size $\gamma_t=K(1+\sum_{i=1}^{t-1}\sum_{k=1}^K\|\hat V_{k,i}-\hat V_{k,i+1/2}\|^2)^{-1/2}$, and quantization levels are updated $J$ times where $\bql_j$ with variance bound $\epsilon_{Q,j}$ in \eqref{varbound} and code-length bound $N_{Q,j}$ in \eqref{codebound} is used for $T_j$ iterations with $\sum_{j=1}^J T_j=T$. Then we have
\begin{align}
\E\Big[\gap_{\Cc}\Big(\frac{1}{T}\sum_{t=1}^T X_{t+1/2}\Big)\Big]=\bigoh\Big(\frac{(\sum_{j=1}^J \sqrt{\epsilon_{Q,j}T_j/T}M+\sigma) D^2}{\sqrt{TK}}\Big)\nn.
\end{align} In addition,~\cref{QGEGalg} requires each processor to send at most $ \frac{2}{T}\sum_{j=1}^J{T_j N_{Q,j}}$ communication bits per iteration in expectation. 
\eth

In the following, we prove the results using the template inequality~\cref{prop:TemIneq} and noise anslysis in~\cref{lm:abs}.

As a preliminary step, we prove this proposition for an {\it adaptive step-size} with a {\it non-adaptive} $Q_\bql$, which satisfies~\cref{thm:varbound}.
\bprop[\cref{QGEGalg} under absolute noise and fixed compression scheme]\label{prop:abs} Under the setup described in~\cref{thm:convQabs}, with an adaptive step-size $\gamma_t=K(1+\sum_{i=1}^{t-1}\sum_{k=1}^K\|\hat V_{k,i}-\hat V_{k,i+1/2}\|^2)^{-1/2}$ and non-adaptive $Q_\bql$ satisfying~\cref{thm:varbound}, we have 
\begin{align}
\E\Big[\gap_{\Cc}\Big(\frac{1}{T}\sum_{t=1}^T X_{t+1/2}\Big)\Big]=\bigoh\Big(\frac{( \sqrt{\epsilon_{Q}}M+\sigma) D^2}{\sqrt{TK}}\Big)\nn.
\end{align}
\eprop
\bpr
Suppose that we do not apply compression, \ie $\epsilon_Q=0$. By the template inequality~\cref{prop:TemIneq}, we have 

\begin{align}\label{eq:Temabs}
\begin{split}
\sum_{t=1}^T\Big\langle\frac{1}{K}\sum_{k=1}^K\hat V_{k,t+1 / 2},X_{t+1 / 2}-X \Big\rangle &\leq\frac{\|X\|_{\ast}^2}{2\g_{T+1}}+\frac{1}{ 2K^2}\sum_{t=1}^T\g_t\sum_{k=1}^K\|\hat V_{k,t+1 / 2}-\hat V_{k,t}\|_{\ast}^2.
\end{split}
\end{align}
Let denote the LHS and RHS of~\eqref{eq:Temabs} by   $S_A=\sum_{t=1}^T\Big\langle\frac{1}{K}\sum_{k=1}^K\hat V_{k,t+1 / 2},X_{t+1 / 2}-X \Big\rangle$ and $S_B=\frac{\|X\|_{\ast}^2}{2\g_{T+1}}+\frac{1}{2K^2}\sum_{t=1}^T\g_t\sum_{k=1}^K\|\hat V_{k,t+1 / 2}-\hat V_{k,t}\|_{\ast}^2$, respectively.Then, by the noise model~\eqref{eq:oracle} and monotonicity of operator $A$, we have 
\begin{align}\nn
\begin{split}
S_A &= \sum_{t=1}^T\Big\langle\frac{1}{K}\sum_{k=1}^K A_{k}(X_{t+1 / 2}),X_{t+1 / 2}-X \Big\rangle+\sum_{t=1}^T\Big\langle\frac{1}{K}\sum_{k=1}^K U_{k,t+1 / 2},X_{t+1 / 2}-X \Big\rangle\\&\geq
\sum_{t=1}^T\Big\langle\frac{1}{K}\sum_{k=1}^K A_{k}(X),X_{t+1 / 2}-X \Big\rangle+\sum_{t=1}^T\Big\langle\frac{1}{K}\sum_{k=1}^K U_{k,t+1 / 2},X_{t+1 / 2}-X \Big\rangle\\&= 
\frac{T}{K}\sum_{k=1}^K\langle A_{k}(X),\ol X_{T+1 / 2}-X\rangle+\sum_{t=1}^T\Big\langle\frac{1}{K}\sum_{k=1}^K U_{k,t+1 / 2},X_{t+1 / 2}-X \Big\rangle
\end{split}
\end{align} where $A_k=A$  for $k\in[K]$. 
Therefore, by rearranging the terms~\cref{eq:Temabs} using above inequality, we have 
\begin{align}\label{eq:Temabs2}
\begin{split}
\frac{T}{K}\sum_{k=1}^K\langle A_{k}(X),\ol X_{T+1 / 2}-X \rangle &\leq - \sum_{t=1}^T\langle\frac{1}{K}\sum_{k=1}^K U_{k,t+1 / 2},X_{t+1 / 2}-X \rangle
\\&\quad+
\frac{\|X\|_{\ast}^2}{2\g_{T+1}}+\frac{1}{ 2K^2}\sum_{t=1}^T\g_t\sum_{k=1}^K\|\hat V_{k,t+1 / 2}-\hat V_{k,t}\|_{\ast}^2.
\end{split}
\end{align}
By taking supermom on both sides of~\cref{eq:Temabs2}, dividing by $T$, and taking expectation, we have 
\begin{align}\label{eq:Temabs3}
\begin{split}
\E\Big[\frac{1}{K}\sum_{k=1}^K\sup_X\langle A_{k}(X),\ol X_{T+1 / 2}-X \rangle\Big] &\leq \frac{1}{T}(S_1+S_2+S_3)\end{split}
\end{align}
where $S_1=\E\l[\frac{D^2}{2\g_{T+1}}\r]$, $S_2=\E\l[\frac{1}{ 2K^2}\sum_{t=1}^T\g_t\sum_{k=1}^K\|\hat V_{k,t+1 / 2}-\hat V_{k,t}\|_{\ast}^2\r]$, and $S_3=\E\l[\sup_X \sum_{t=1}^T\langle\frac{1}{K}\sum_{k=1}^K U_{k,t+1 / 2},X_{t+1 / 2}-X \rangle\r]$. 
We now bound $S_1$, $S_2$, and $S_3$ from above, individually.  
For $S_1$, we have 
\begin{align}\label{eq:S1}
\begin{split}
S_1&=\E\l[\frac{D^2}{2\g_{T+1}}\r]\\
&=\frac{D^2}{2K}\E\l[\sqrt{1+\sum_{i=1}^{T-1}\sum_{k=1}^K\|\hat V_{k,i}-\hat V_{k,i+1/2}\|^2}\r]\\
&\leq\frac{D^2}{2K}\sqrt{1+\sum_{i=1}^{T-1}\sum_{k=1}^K\E\l[\|\hat V_{k,i}-\hat V_{k,i+1/2}\|^2\r]}\\
&\leq \frac{D^2}{2K}\sqrt{1+\sum_{i=1}^{T-1}\sum_{k=1}^K2(\E[\|\hat V_{k,i}\|^2]+\E[\|\hat V_{k,i+1/2}\|^2])}\\
&\leq\frac{D^2}{2K}\sqrt{1+4KT\s^2}. 
\end{split}    
\end{align}
We also have 
\begin{align}\small\label{eq:S2}
\begin{split}
S_2&=\E\l[\frac{1}{2K^2}\sum_{t=1}^T\g_t\sum_{k=1}^K\|\hat V_{k,t+1 / 2}-\hat V_{k,t}\|_{\ast}^2\r]\\
&=\frac{1}{2}\E\l[\sum_{t=1}^T
\l(\frac{\g_t}{K^2}-\frac{\g_{t+1}}{K^2}\r)\sum_{k=1}^K\|\hat V_{k,t+1 / 2}-\hat V_{k,t}\|_{\ast}^2\r]+\frac{1}{2}\E\l[\sum_{t=1}^T\frac{\g_{t+1}}{K^2}\sum_{k=1}^K\|\hat V_{k,t+1 / 2}-\hat V_{k,t}\|_{\ast}^2\r]\\
&\leq 2\E\l[\sum_{t=1}^T
\l(\frac{\g_t}{K^2}-\frac{\g_{t+1}}{K^2}\r)K\s^2\r]+\frac{1}{2}\E\l[\sum_{t=1}^T\frac{\g_{t+1}}{K^2}\sum_{k=1}^K\|\hat V_{k,t+1 / 2}-\hat V_{k,t}\|_{\ast}^2\r]\\
&\leq 2\s^2+\frac{1}{2K}\E\l[
\sum_{t=1}^T\frac{\sum_{k=1}^K\|\hat V_{k,t+1 / 2}-\hat V_{k,t}\|_{\ast}^2}{\sqrt{1+\sum_{t=1}^T\sum_{k=1}^K\|\hat V_{k,t+1 / 2}-\hat V_{k,t}\|_{\ast}^2}}\r]\\
&\leq 2\s^2+\frac{1}{2K}\E\l[
\sqrt{1+\sum_{t=1}^T\sum_{k=1}^K\|\hat V_{k,t+1 / 2}-\hat V_{k,t}\|_{\ast}^2}\r]\\
&\leq 2\s^2+\frac{1}{2K}
\sqrt{1+\sum_{t=1}^T\sum_{k=1}^K\E\l[\|\hat V_{k,t+1 / 2}-\hat V_{k,t}\|_{\ast}^2\r]}\\
&\leq 2\s^2+\frac{1}{2K}\sqrt{1+4\s^2KT}.
\end{split}    
\end{align}

Finally we note that 

\begin{align}\label{eq:S3}
\begin{split}
S_3&=\E\l[\sup_X \sum_{t=1}^T\langle\frac{1}{K}\sum_{k=1}^K U_{k,t+1 / 2},X_{t+1 / 2}-X \rangle\r]\\&=\E\l[\sup_X \sum_{t=1}^T\langle\frac{1}{K}\sum_{k=1}^K U_{k,t+1 / 2},X \rangle\r]-\E\l[\sup_X \sum_{t=1}^T\langle\frac{1}{K}\sum_{k=1}^K U_{k,t+1 / 2},X_{t+1 / 2} \rangle\r].
\end{split}    
\end{align}

We bound the first term in the RHS of~\eqref{eq:S3} using the following known lemma: 
\blm[{\citealt{BL19}}]\label{lm:S3}
Let $\Cc\in\reals^d$ be a convex set and $h:\Cc\ra\reals$ be a 1-strongly convex w.r.t. a norm $\|\cdot\|$. Assume that $h(\xbf)-\min_{\xbf\in\Cc} h(\xbf)\leq D^2/2$ for all $\xbf\in\Cc$. Then, for any martingle difference $(\zbf_t)_{t=1}^T\in\reals^d$ and any  $\xbf\in\Cc$, we have 
\begin{align}\nn
\E\l[\Big\langle\sum_{t=1}^T\zbf_t,\xbf\Big\rangle\r]\leq\frac{D^2}{2}\sqrt{\sum_{t=1}^T\E[\|\zbf_t\|^2]}.     
\end{align}
\elm

Using~\cref{lm:S3}, the first term  in the RHS of~\eqref{eq:S3} is bounded by

\begin{align}\label{eq:S3bound}
\begin{split}
\frac{1}{K}\E\l[\sup_X \sum_{t=1}^T\langle\sum_{k=1}^K U_{k,t+1 / 2},X \rangle\r]&\leq
\frac{D^2}{2K}\sqrt{\E\l[\sum_{t=1}^T\sum_{k=1}^K \|U_{k,t+1 / 2}\|^2\r]}\\&\leq
\frac{D^2\s\sqrt{\nRuns}}{2\sqrt{K}}.
\end{split}    
\end{align}

Similarly, we can bound the second term  in the RHS of~\eqref{eq:S3}. Combining the results in~\cref{eq:S1}, \cref{eq:S2}, and \cref{eq:S3bound} and applying ~\cref{lm:abs}, we obtain the upper bound in~\cref{prop:abs}. 
\epr 

Applying~\cref{prop:abs} with the scaled step size schedule in~\cref{thm:convQabs}, we complete the proof for an adaptive compression scheme along the lines of~\citep[Theorem 4]{ALQ}. 

\section{Proof of \cref{thm:convQrel} (\ref{eq:QGEG} under relative noise)}\label{app:convQrel}

We first remind the theorem statement: 
\bth[\ref{eq:QGEG} under relative noise]%
Let $\Cc\subset\reals^d$ denote a compact neighborhood of a solution for \eqref{eq:VI} and let %
$D^2:=\sup_{X\in\Cc} \|X-X_0\|^2$. Suppose that the oracle and the problem~\eqref{eq:VI} satisfy Assumptions~\ref{assumption:ex},~\ref{assumption:relative}, and \ref{assumption:co},~\cref{QGEGalg}  
is executed for $T$ iterations on $K$ processors 
with an adaptive step-size $\gamma_t=K(1+\sum_{i=1}^{t-1}\sum_{k=1}^K\|\hat V_{k,i}-\hat V_{k,i+1/2}\|^2)^{-1/2}$,
and quantization levels are updated $J$ times where $\bql_j$ with variance bound $\epsilon_{Q,j}$ in \eqref{varbound} and code-length bound $N_{Q,j}$ in \eqref{codebound} is used for $T_j$ iterations with $\sum_{j=1}^J T_j=T$. Then we have
\begin{align}
\E\Big[\gap_{\Cc}\Big(\frac{1}{T}\sum_{t=1}^T X_{t+1/2}\Big)\Big]=\bigoh\Big(\frac{\big((c+1)\sum_{j=1}^JT_j \epsilon_{Q,j}/T+c\big) D^2}{KT}\Big).\nn
\end{align} In addition,~\cref{QGEGalg} requires each processor to send at most $ \frac{2}{T}\sum_{j=1}^J{T_j N_{Q,j}}$ communication bits per iteration in expectation. 
\eth
Suppose that we do not apply compression, \ie $\epsilon_Q=0$. We first remind the template inequality in~\eqref{eq:TemIneq}, which holds for any $\g_t$ and noise model: 

\begin{align}\nn%
\begin{split}
\sum_{t=1}^T\Big\langle\frac{1}{K}\sum_{k=1}^K\hat V_{k,t+1 / 2},X_{t+1 / 2}-X \Big\rangle &\leq\frac{\|X\|_{\ast}^2}{2\g_{T+1}}+\frac{1}{ 2K^2}\sum_{t=1}^T\g_t\sum_{k=1}^K\|\hat V_{k,t+1 / 2}-\hat V_{k,t}\|_{\ast}^2\\&\quad -\frac{1}{2}\sum_{t=1}^T\frac{1}{\g_t}\|X_t-X_{t+1 / 2}\|_{\ast}^2.    
\end{split}
\end{align}

In the following proposition, we show that under $\g_t$ and relative noise model in~\cref{thm:convQrel}, 
$\sum_{t=1}^T\E\big[\| A(X_{t+1 / 2})\|_{\ast}^2+\| A(X_{t})\|_{\ast}^2\big]$ is summable in the sense that $\sum_{t=1}^T\E\big[\| A(X_{t+1 / 2})\|_{\ast}^2+\| A(X_{t})\|_{\ast}^2\big]=\bigoh(1/\g_{T})$.

\bprop[Sum operator output under relative noise]\label{prop:sumop} Let $X^*$ denote a solution of~\eqref{eq:VI}. Under the setup described in~\cref{thm:convQrel}, we have: 
\begin{align}\label{eq:sumop}\sum_{t=1}^T\E\big[\| A(X_{t+1 / 2})\|_{\ast}^2+\| A(X_{t})\|_{\ast}^2\big]\leq\E\Big[\frac{\|X^*\|_{\ast}^2}{2\g_{T+1}}\Big].
\end{align}
\eprop
\bpr Substituting $X=X^*$ into  $\E\big[\big\langle\frac{1}{K}\sum_{k=1}^K\hat V_{k,t+1 / 2},X_{t+1 / 2}-X \big\rangle\big]$ and applying the law of total expectation, we have: 
\begin{align}\label{eq:Xstar}
\begin{split}
\E\Big[\Big\langle\frac{1}{K}\sum_{k=1}^K\hat V_{k,t+1 / 2},X_{t+1 / 2}-X^* \Big\rangle\Big] &=\E\Big[\frac{1}{K}\sum_{k=1}^K\E[\langle\hat V_{k,t+1 / 2},X_{t+1 / 2}-X^*\rangle | X_{t+1 / 2}]\Big]\\
&=\E\Big[\frac{1}{K}\sum_{k=1}^K\langle A_{k}(X_{t+1 / 2}),X_{t+1 / 2}-X^*\rangle \Big]\\
&=\E\Big[\langle A(X_{t+1 / 2}),X_{t+1 / 2}-X^*\rangle \Big]\\
&\geq\E\Big[\langle A(X_{t+1 / 2})-A(X^*),X_{t+1 / 2}-X^*\rangle\Big]\\
&\geq\smooth\E[\| A(X_{t+1 / 2})\|_{\ast}^2]
\end{split}
\end{align} where the fourth and fifth inequalities hold due to the definition of the monotone operator~\cref{eq:VI} and $\smooth$-cocoecivity in~\cref{eq:coco}, respectively.

Applying the lower bound in~\cref{eq:Xstar} into~\cref{eq:TemIneq}, we obtain: 
\begin{align}\label{eq:TemIneqXstar}
\begin{split}
\sum_{t=1}^T\smooth\E[\| A(X_{t+1 / 2})\|_{\ast}^2] &\leq\E\Big[\frac{\|X^*\|_{\ast}^2}{2\g_{T+1}}+\frac{1}{ 2K^2}\sum_{t=1}^T\g_t\sum_{k=1}^K\|\hat V_{k,t+1 / 2}-\hat V_{k,t}\|_{\ast}^2\\&\quad -\frac{1}{2}\sum_{t=1}^T\frac{1}{\g_t}\|X_t-X_{t+1 / 2}\|_{\ast}^2\Big].    
\end{split}
\end{align}
Moreover, by lower bounding (LHS) of the above:
\begin{align}
    \sum_{t=1}^{T}\E[\|A(X_{t+1/2})\|^{2}_{\ast}&= \sum_{t=1}^{T}\E[1/K\sum_{k=1}^{K}\|
    A(X_{t+1/2})\|^{2}]\\
    &\geq \frac{1}{c}\sum_{t=1}^{T}\E[1/K\sum_{k=1}^{K}\|\hat V_{k,t+1 / 2}\|^{2}_{\ast}]
\end{align}
with the second inequality being obtained by the relative noise condition.
On the other hand, applying Cauchy–Schwarz and $\smooth$-cocoecivity in~\cref{eq:coco} imply $\|X_t-X_{t+1 / 2}\|_{\ast}^2\geq \smooth^2\|A(X_t)-A(X_{t+1 / 2})\|_{\ast}^2$. It follows that: 
\begin{align}\label{eq:Xstarharlf}
\begin{split}
&\frac{1}{2}\E\Big[\sum_{t=1}^T\smooth\| A(X_{t+1 / 2})\|_{\ast}^2+\sum_{t=1}^T\frac{1}{\g_t}\|X_t-X_{t+1 / 2}\|_{\ast}^2\Big]\\& \geq\frac{1}{2}\E\Big[\sum_{t=1}^T\smooth\| A(X_{t+1 / 2})\|_{\ast}^2+\sum_{t=1}^T\frac{\smooth^2}{\g_t}\|A(X_t)-A(X_{t+1 / 2})\|_{\ast}^2\Big]\\& \geq\frac{1}{2}\E\Big[\sum_{t=1}^T\smooth\| A(X_{t+1 / 2})\|_{\ast}^2+\sum_{t=1}^T\frac{\smooth^2}{\g_t}\|A(X_t)-A(X_{t+1 / 2})\|_{\ast}^2\Big]\\& \geq\frac{1}{2}\min\Big\{\smooth,\frac{\smooth^2}{\g_0}\Big\}\sum_{t=1}^T\E\Big[\| A(X_{t+1 / 2})\|_{\ast}^2+\|A(X_t)-A(X_{t+1 / 2})\|_{\ast}^2\Big]\\&
\geq\frac{1}{2}\min\Big\{\smooth,\frac{\smooth^2}{\g_0}\Big\}\sum_{t=1}^T\E\Big[\| A(X_{t})\|_{\ast}^2\Big]\\& 
\geq\frac{1}{2}\min\Big\{\smooth,\frac{\smooth^2}{\g_0}\Big\}\sum_{t=1}^{T}\E\Big[1/K\sum_{k=1}^{K}\|A(X_{t})\|^{2}_{\ast}\Big]\\&
\geq\frac{1}{2c}\min\Big\{\smooth,\frac{\smooth^2}{\g_0}\Big\}\sum_{t=1}^{T}\E\Big[1/K\sum_{k=1}^{K}\|\hat V_{k,t}\|^{2}_{\ast}\Big]
\end{split}
\end{align} where the last inequality holds due to the relative noise condition.
Combining the above inequalities we get the following:
\begin{equation}
 \frac{\beta}{c}\sum_{t=1}^{T}\E[1/K\sum_{k=1}^{K}\|\hat V_{k,t+1 / 
 2}\|^{2}_{\ast}]\leq \E[\frac{\|X\|_{\ast}^2}{2\g_{T+1}}+\frac{1}{ 
 2K^2}\sum_{t=1}^T\g_t\sum_{k=1}^K\|\hat V_{k,t+1 / 2}-\hat 
 V_{k,t}\|_{\ast}^2 ]     
 \end{equation}
 and
 \begin{equation}
  \frac{1}{2c}\min\Big\{\smooth,\frac{\smooth^2}{\g_0}\Big\}\sum_{t=1}^{T}\E\Big[1/K\sum_{k=1}^{K}\|\hat V_{k,t}\|^{2}_{\ast}\Big] \leq \E[\frac{\|X\|_{\ast}^2}{2\g_{T+1}}+\frac{1}{ 2K^2}\sum_{t=1}^T\g_t\sum_{k=1}^K\|\hat V_{k,t+1 / 2}-\hat V_{k,t}\|_{\ast}^2]       
 \end{equation}
Therefore, by adding the above inequalities we get:
\begin{align}\label{sumVs}
\begin{split}
 &\frac{\beta}{c}\sum_{t=1}^{T}\E[1/K\sum_{k=1}^{K}\|\hat V_{k,t+1 /2}\|^{2}_{\ast}]+ \frac{1}{2c}\min\Big\{\smooth,\frac{\smooth^2}{\g_0}\Big\}\sum_{t=1}^{T}\E\Big[1/K\sum_{k=1}^{K}\|\hat V_{k,t}\|^{2}_{\ast}\Big]\\&\leq 2\E
 \Big[\frac{\|X\|_{\ast}^2}{2\g_{T+1}}
 +\frac{1}{ 2K^2}\sum_{t=1}^T\g_t\sum_{k=1}^K\|\hat V_{k,t+1 / 2}-\hat V_{k,t}\|_{\ast}^2 \Big]
 \end{split}
\end{align}
We now establish an upper bound on the R.H.S. of~\cref{sumVs}. 
We first note that:
\begin{align}\label{sumVsUBstep}
\begin{split}
 &\E\big[\frac{1}{ 2K^2}\sum_{t=1}^T\g_t\sum_{k=1}^K\|\hat V_{k,t+1 / 2}-\hat V_{k,t}\|_{\ast}^2\big]\\&=\E\big[ \frac{1}{ 2K^2}\sum_{t=1}^T(\g_t-\g_{t+1})\sum_{k=1}^K\|\hat V_{k,t+1 / 2}-\hat V_{k,t}\|_{\ast}^2
 +\frac{1}{ 2K^2}\sum_{t=1}^T\g_{t+1}\sum_{k=1}^K\|\hat V_{k,t+1 / 2}-\hat V_{k,t}\|_{\ast}^2\big]\\&\leq
\E\big[\frac{1}{2K}(4M^2)\sum_{t=1}^T(\g_t-\g_{t+1})+\frac{1}{ 2K^2}\sum_{t=1}^T\g_{t+1}\sum_{k=1}^K\|\hat V_{k,t+1 / 2}-\hat V_{k,t}\|_{\ast}^2\big]\\&\leq  
\E\big[\frac{2M^2}{K}\g_1+\frac{1}{ 2K^2}\sum_{t=1}^T\g_{t+1}\sum_{k=1}^K\|\hat V_{k,t+1 / 2}-\hat V_{k,t}\|_{\ast}^2\big]\\&\leq 
\E\big[{2M^2}+\frac{1}{ 2K^2}\sum_{t=1}^T\g_{t+1}\sum_{k=1}^K\|\hat V_{k,t+1 / 2}-\hat V_{k,t}\|_{\ast}^2\big]\\&\leq 
\E\Big[{2M^2}+\frac{1}{ K}\sqrt{1+\sum_{t=1}^T\sum_{k=1}^K\|\hat V_{k,t+1 / 2}-\hat V_{k,t}\|_{\ast}^2}\Big]\\&\lesssim 
\E[\frac{1}{\g_{T+1}}].
\end{split}
\end{align}

Therefore, an upper bound on the R.H.S. of~\cref{sumVs} is given by :
\begin{align}\label{sumVsUB}
\begin{split}
\frac{\beta}{c}\sum_{t=1}^{T}\E[1/K\sum_{k=1}^{K}\|\hat V_{k,t+1 /2}\|^{2}_{\ast}]+ \frac{1}{2c}\min\Big\{\smooth,\frac{\smooth^2}{\g_0}\Big\}\sum_{t=1}^{T}\E\Big[1/K\sum_{k=1}^{K}\|\hat V_{k,t}\|^{2}_{\ast}\Big]\leq \E
 \Big[\frac{\|X^{\ast}\|_{\ast}^2+1}{\g_{T+1}} \Big].
 \end{split}
\end{align}
\epr
To establish a lower bound on the  L.H.S. of~\cref{sumVs}, we first note that: 

\begin{align}\label{sumVsLBstep}
\begin{split}
\E\Big[\frac{1}{ K^2}\sum_{t=1}^T\sum_{k=1}^K\|\hat V_{k,t+1 / 2}-\hat V_{k,t}\|_{\ast}^2\Big]&=\E\Big[\frac{1}{ K^2}\Big(1+\sum_{t=1}^T\sum_{k=1}^K\|\hat V_{k,t+1 / 2}-\hat V_{k,t}\|_{\ast}^2\Big)\Big]-\frac{1}{K^2}\\
&= \E[\frac{1}{\g_{T+1}^2}]-\frac{1}{K^2}.
\end{split}
\end{align}

We also note that: 

\begin{align}\label{sumLBstep}
\begin{split}
&\frac{K}{4c}\min\Big\{\smooth,\frac{\smooth^2}{\g_0}\Big\}\E\big[\big(1/K^2\sum_{t=1}^{T}\sum_{k=1}^{K}\|\hat V_{k,t+1 /2}-\hat V_{k,t}\|^{2}_{\ast}\big)\big]\\&\leq \frac{\beta}{2c}\sum_{t=1}^{T}\E[1/K\sum_{k=1}^{K}\|\hat V_{k,t+1 /2}\|^{2}_{\ast}]+ \frac{1}{2c}\min\Big\{\smooth,\frac{\smooth^2}{\g_0}\Big\}\sum_{t=1}^{T}\E\Big[1/K\sum_{k=1}^{K}\|\hat V_{k,t}\|^{2}_{\ast}\Big].
 \end{split}
\end{align}

Hence, combining,~\cref{sumVsUB,sumVsLBstep,sumLBstep}, we can find an upper bound on $\E[\frac{1}{\g_{T+1}^{2}}]
$:
\begin{align*}
 \frac{K}{4c}\min\Big\{\smooth,\frac{\smooth^2}{\g_0}\Big\}\E[\frac{1}{\g_{T+1}^{2}}]
 &\leq \big[\|X^{\ast}\|^{2}+1\big]\E[\frac{1}{\g_{T+1}}]
 \\
 &=\big[\|X^{\ast}\|^{2}+1\big]\E[\sqrt{\frac{1}{\g_{T+1}^{2}}}]
\\
&\leq \big[\|X^{\ast}\|^{2}+1\big]\sqrt{\E[\frac{1}{\g_{T+1}^{2}}]}
\end{align*}
with the last inequality being obtained by Jensen's inequality. So we have 
\begin{align}\label{gammaT1UB}
 \E[\frac{1}{\g_{T+1}}]
 \leq \frac{4c}{K}\max\Big\{\frac{1}{\smooth},\frac{\g_0}{\smooth^2}\Big\}.
\end{align}

Similar to the proof of  \cref{thm:convQabs}, we have 

\begin{align}\label{eq:s3rel}
\begin{split}
\E\Big[\sup_X\langle A(X),\ol X_{T+1 / 2}-X \rangle\Big] &\leq \frac{1}{T}(S_1+S_2+S_3)\end{split}
\end{align}
where $S_1=\E\l[\frac{D^2}{2\g_{T+1}}\r]$, $S_2=\E\l[\frac{1}{ 2K^2}\sum_{t=1}^T\g_t\sum_{k=1}^K\|\hat V_{k,t+1 / 2}-\hat V_{k,t}\|_{\ast}^2\r]$, and $S_3=\E\l[\sup_X \sum_{t=1}^T\langle\frac{1}{K}\sum_{k=1}^K U_{k,t+1 / 2},X-X_{t+1 / 2} \rangle\r]$.

By~\eqref{sumVsUBstep}, we have 

\begin{align}\label{eq:s2rel}
\begin{split}
\E\l[\frac{1}{ 2K^2}\sum_{t=1}^T\g_t\sum_{k=1}^K\|\hat V_{k,t+1 / 2}-\hat V_{k,t}\|_{\ast}^2\r] &\lesssim\E\Big[\frac{1}{\g_{T+1}}\Big].
\end{split}
\end{align}

We now decompose $S_3$ into two terms $S_3=\E\l[\sup_X \sum_{t=1}^T\langle\frac{1}{K}\sum_{k=1}^K U_{k,t+1 / 2},X \rangle\r]-\E\l[ \sum_{t=1}^T\langle\frac{1}{K}\sum_{k=1}^K U_{k,t+1 / 2},X_{t+1 / 2} \rangle\r]$.

Let the supremum on the first terms is attained by $X=X^o$. We can also establish an upper bound on the first term using~\cref{lm:S3}: 

\begin{align}\label{eq:s3t1}
\begin{split}
\frac{1}{K}\E\l[\sup_X \sum_{t=1}^T\langle\sum_{k=1}^K U_{k,t+1 / 2},X \rangle\r]&=\frac{1}{K}\E\l[ \langle\sum_{t=1}^T\sum_{k=1}^K U_{k,t+1 / 2},X^o \rangle\r]\\&
=\frac{D^2}{2K}\sqrt{\E\Big[ \|\sum_{t=1}^T\sum_{k=1}^K U_{k,t+1 / 2}\|_{\ast}^2\Big]}\\&
\leq \frac{D^2}{2\sqrt{K}}\sqrt{\E\Big[\sum_{t=1}^Tc\| A(X_{t+1 / 2})\|_{\ast}^2\Big]}\\&
\leq \frac{D^2}{2\sqrt{K}}\sqrt{c\E\Big[\frac{\|X^*\|_{\ast}^2}{2\g_{T+1}}\Big]}
\end{split}
\end{align} where the last inequality holds by~\cref{prop:sumop}. 

Finally, by the law of total expectation, we have: 

\begin{align}\label{eq:s3t2}
\begin{split}
\E\l[\sum_{t=1}^T\langle\sum_{k=1}^K U_{k,t+1 / 2},X_{t+1 / 2}  \rangle\r]&=\E\l[ \sum_{t=1}^T\sum_{k=1}^K \E[\langle U_{k,t+1 / 2},X_{t+1 / 2}  \rangle| X_{t+1 / 2}]\r]=0.
\end{split}
\end{align}

Substituting~\cref{gammaT1UB}, \cref{eq:s3t1}, and~\cref{eq:s3t2} into ~\cref{eq:s3rel}, we have 

\begin{align}\label{gaprelstar}
\E[\gap_{\Cc}\ol X_{T+1 / 2}]=\bigoh\Big(\frac{1}{T}\E\Big[\frac{D^2}{2\g_{T+1}}\Big]\Big).
\end{align}

Following similar analysis as in~\cref{prop:abs} and applying~\cref{lm:rel}  with the scaled step size schedule in~\cref{thm:convQrel}, we complete the proof for an adaptive compression scheme along the lines of~\citep[Theorem 4]{ALQ}.

\section{Experimental details and additional experiments}
\label{app:experiment}

In order to validate our theoretical results, we build on the code base of \citet{GBVV+19} and run an instantiation of Q-GenX obtained by combining ExtraAdam with the compression offered by the \texttt{torch\_cgx} pytorch extension of~\citet{CGX}, and train a WGAN-GP~\citep{arjovsky2017wasserstein} on CIFAR10~\citep{CIFAR10}.

\begin{figure}
    \centering
    \begin{subfigure}[b]{0.5\textwidth}
    \includegraphics[width=\textwidth]{figures/FID_evolution.pdf}
    \caption{FID evolution during training}
    \label{figapp:fid_vs_time}
    \end{subfigure}%
    \begin{subfigure}[b]{0.5\textwidth}
    \includegraphics[width=\textwidth]{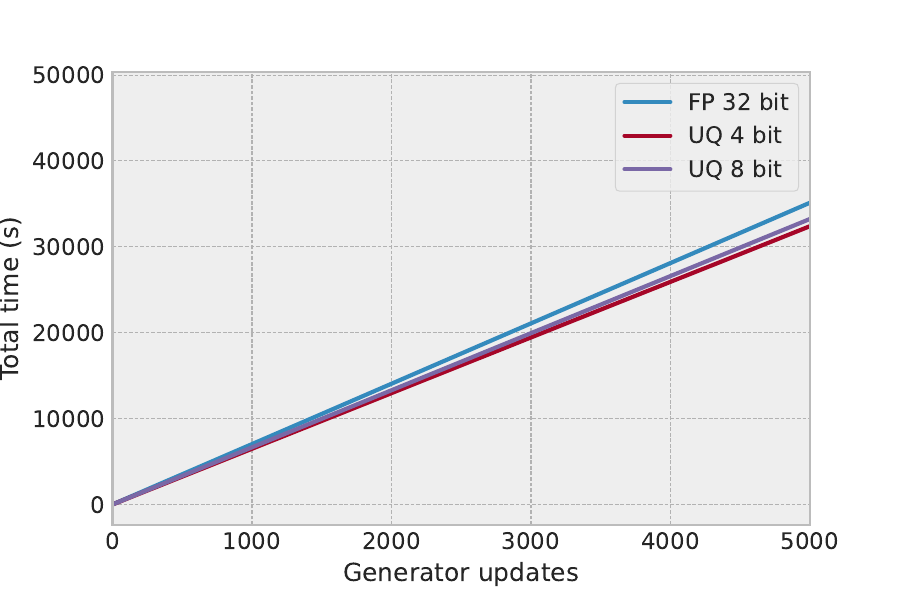}
    \caption{Total time spent on backpropagation and gradient exchanges}
    \label{figapp:wall_clock}
    \end{subfigure}
    \label{figapp:comparison}
    \caption{Comparing full gradient ExtraAdam with a simple instantiation of QGenX. FID stands for Frechet inception distance, which is a standard GAN quality metric introduced in~\citep{FID}.}
\end{figure}

\begin{figure}
\centering
\begin{minipage}{0.5\textwidth}
\centering
    \includegraphics[width=\textwidth]{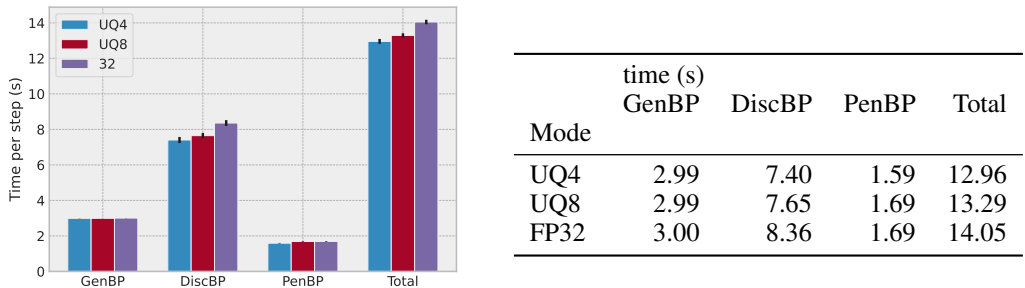}
    \label{figapp:fine_grained}
\end{minipage}%
\begin{minipage}{0.5\textwidth}
\centering
\captionsetup{type=table} %
\begin{tabular}{lrrrr}
\toprule
{} & \multicolumn{4}{l}{time (s)} \\
&     GenBP &    DiscBP &     PenBP &      Total \\
Mode  &           &           &           &            \\
\midrule
UQ4 & 2.99  & 7.40 & 1.59 & 12.96 \\
UQ8 & 2.99  & 7.65  & 1.69  & 13.29 \\
FP32 & 3.00  & 8.36  & 1.69  & 14.05 \\
\bottomrule
\end{tabular}
\label{tabapp:fine_grained}
\end{minipage}
    \caption{Fine grained comparison of average \texttt{.backward()} times  on generator, discriminator, gradient penalty as well as total training time. The \texttt{.backward()} function is where pytorch DDP handles gradient exchange}
    \label{fine_grainedapp}
\end{figure}

Since \texttt{torch\_cgx} uses OpenMPI~\citep{gabriel04:_open_mpi} as its communication backend, we use OpenMPI  as the communication backend for the full gradient as well for a fairer comparison.
We deliberately do \emph{not} tune any hyperparameters to fit the larger batchsize since simliar to~\citep{GBVV+19}, we do not claim to set a new SOTA with these experiments but simply want to show that our theory holds up in practice and can potentially lead to improvements. For this, we present a basic experiment showing that even for a {\it very small problem size} and  {\it a heuristic base compression method of cgx}, we can achieve a {\it noticeable speedup} of around $8\%$. We expect further gains to be achievable for larger problems and more advanced compression methods. Given differences in terms of settings and the {\it lack of any code, let alone an efficient implementations that can be used in a real-world setting (i.e. CUDA kernels integrated with networking)}, it is difficult to impossible to conduct a fair comparison with \citet{beznosikov2021distributed}. %

We follow exactly the setup of \citep{GBVV+19} except that we share an effective batch size of $1024$ across 3 nodes (strong scaling) connected via Ethernet, and use Layernorm~\citep{ba2016layer} instead of Batchnorm~\citep{ioffe2015batch} since Batchnorm is known to be challenging to work with in distributed training as well as interacting badly with the WGAN-GP penalty. The results are shown in \cref{figapp:fid_vs_time}, showing evolution of FID and \cref{figapp:wall_clock} showing the accumulated total time spent backpropagating.
We note that we do \emph{not} scale the learning rate or any other hyperparameters to account for these two changes so this experiment is \emph{not} meant to claim SOTA performance, merely to illustrate that 
\begin{enumerate}
    \item Even with the simplest possible unbiased quantization on a relatively small-scale setup, we can observe speedup (about $10\%$).
    \item This speedup does not drastically change the performance.
\end{enumerate}
We compare training using the full gradient  of 32 bit (FP32) to training with gradients compressed to $8$ (UQ8) and $4$ bits  (UQ4) using a bucket size of $1024$. \Cref{fine_grainedapp} shows a more fine grained breakdown of the time used for back propagation (BP)  where the network activity takes place. GenBP, DiscBP and PenBP refer to the backpropagation for generator, discriminator and the calculation of the gradient penalty, respectively. Total refers to the sum of these times.\\ 

We used Weights and Biases~\citep{wandb} for all experiment tracking. Our time measurements are performed with pythons \texttt{time.time()} function which has microsecond precision on Linux, measuring only backward propagation times and total training time, excluding plotting, logging etc.
The experiments were performed on 3 Nvidia V100 GPUs (1 per node) using a Kubernetes cluster and an image built on the \texttt{torch\_cgx} Docker image.
\subsection{Comparison with QSGDA}
\begin{figure}
    \centering
    \caption{Comparing Q-GenX with QSGDA }
    \includegraphics[width=\textwidth]{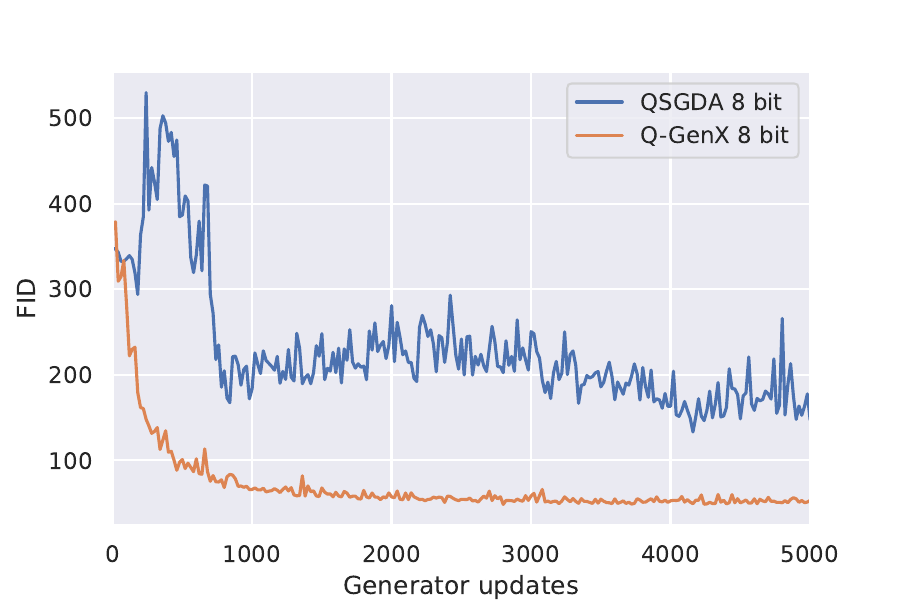}
    \label{figapp:q_sgda}
\end{figure}
\Cref{figapp:q_sgda} compares Q-GenX with QSGDA of \citet{beznosikov2022unifiedtheory}, the only method without explicit variance reduction.
Due to the extra-gradient template, Q-GenX is able to make steady progress without variance reduction.

\section{Trade-off between number of iterations and time per iteration}\label{app:tradeoff}
In this section, we build on our theoretical results in \cref{thm:convQabs,thm:convQrel} to capture the trade-off between the number of iterations to converge and time per iteration, which includes total time required to update a model on each GPU. 

Using the results of~\cref{thm:convQabs,thm:convQrel}, we can obtain the minimum number of iterations required to {\it guarantee an expected gap of $\epsilon$}, which is a measure of solution quality. In particular, under absolute noise model and average gradient variance bound $\ol\epsilon_Q=\sum_{j=1}^JT_j \epsilon_{Q,j}/T$, the minimum number of iterations to guarantee  an expected gap of $\epsilon$ is $T(\epsilon,\ol\epsilon_Q)= (\ol\epsilon_QM^2+\sigma^2)^2D^4/\epsilon^2$. Now suppose that the time per iteration, which includes overall computation, encoding, and communication times to compute, compress, send, receive, decompress and update one iteration, is denoted by $\Delta$. Decreasing the number of communication bits, i.e., compressing more aggressively increases  the sufficient number of iterations $T(\epsilon,\ol\epsilon_Q)$ and decreases time per iteration, $\Delta$, due to communication savings, which captures the trade-off. Theoretically, the best compression method is the one with the minimum overall wall-clock time bounded by $T(\epsilon,\ol\epsilon_Q)\Delta$. The exact optimal point depends on the specific problem to solve (dataset, loss, etc.) and the hyperparameters chosen (architecture, number of bits, etc.), which together determine $\ol\epsilon_Q$ and the implementation details of the algorithm, networking, and compression, along with  the  cluster setup and hyperparameters, which together influence $\Delta$. We defer more refined analysis of the optimal point to future work.

\section{Examples motivating \cref{assumption:relative}}\label{app:example}
In this section, we provide some popular examples which motivate \cref{assumption:relative}:
\begin{example}
[Random coordinate descent (RCD)]
Consider a smooth convex function $\obj$ over $\reals^{\vdim}$.
At iteration $t$, the RCD algorithm draws one coordinate $i_t\in[d]$ uniformly at random and computes the partial derivative $v_{i,t} = \partial f / \partial x_{i_t}$. Subsequently, the $i$-th derivative is updated as $X_{i,t+1} = X_{i,t} - d\cdot\a\cdot v_{i,t}$ where $\a>0$ denotes the step-size.

This update rule can be written in an abstract recursive form as
$\xbf^{+} = \xbf -\a g(\xbf;\mu)$
where $g_{i}(\xbf;\mu) = \vdim \cdot \partial \obj / \partial x_i  \cdot \mu$ and $\mu$ is drawn uniformly at random from the set of basis vectors $\{\ebf_1,\dotsc,\ebf_d\}\subseteq\mathbb{R}^{\vdim}$.
We note that $\exof{g(\xbf;\mu)} = \nabla\obj(\xbf)$. Furthermore, since $\partial f / \partial x_i = 0$ at the minima of $f$, we also have $g(\xbf^{*};\mu) = 0$ if $\xbf^{*}$ is a minimizer of $f$, \ie the variance of the random vector $g(\xbf;\mu)$ vanishes at the minima of $\obj$.
It is not difficult to show that $\E_{\mu}{\|g(\xbf;\mu) - \nabla f(\xbf)\|^2} = \bigoh(\|\nabla f(x)\|^2)$, which satisfies~\cref{assumption:relative} with $A = \nabla\obj$. 
\end{example}

\begin{example}
[Random player updating] Consider an $N$-player convex game with loss functions $f_i$, $i\in[N]$. Suppose, at each stage, player $i$ is selected with probability $p_i$ to play an action following its individual gradient descent rule $X_{i,t+1} = X_{i,t} + \gamma_{t}/p_{i} V_{i,t}$
where $V_{i,t} = \nabla_i f_i(X_t)$ denotes player $i$'s individual gradient at the state $X_t = (X_{1,t},...,X_{N,t})$ and $p_i$ is included for scaling reasons.

Note that $\mathbb{E}[V_t] = A(X_t)$ where $A_i(x) = \nabla_i f_i(x)$ for $i\in[N]$. It is not difficult to show that $V_t$ is an {\it unbiased oracle} for $A$, and since all individual components of $A$ vanish at the game's Nash equilibria, it is also straightforward to verify that $V_t$ satisfies~\cref{assumption:relative}.
\end{example}

\section{Encoding}\label{app:encoding} To further reduce communication costs, we can apply information-theoretically inspired coding schemes on top of quantization. In this section, we provide an overview of our coding schemes along the lines of~\citep{QSGD,ALQ,NUQSGD}. Let $q\in\integers_+$. We first note that a vector $\vbf\in\reals^d$ can be {\it uniquely} represented by a tuple  $(\|\vbf\|_q,\signvec,\ubf)$
where $\|\vbf\|_q$ is the $L^q$ norm of $\vbf$,
$\signvec:=[\sign(v_1),\ldots,\sign(v_d)]^\top$ consists of signs of the coordinates $v_i$'s,
and $\ubf:=[u_1,\ldots,u_d]^\top$ with $u_i=|v_i|/\|\vbf\|_q$ are the normalized coordinates. Note that $0 \leq u_i\leq 1$ for all $i\in[d]$. We  define a random quantization function as follows: 
\begin{definition}[Random quantization  function]\label{defapp:quantf}
Let $s\in\integers_+$ denote the number of quantization levels. Let $u\in[0,1]$ and $\bql= (\ql_0,\ldots,\ql_{s+1})$ denote a sequence of $s$ {\it quantization levels} with $0=\ql_0<\ql_1<\cdots<\ql_s<\ql_{s+1}=1$. Let ${\level{u}}$ denote the index of a level such that 
${\ql_{\level{u}}\leq u<\ql_{\level{u}+1}}$.
Let $\qcoeff(u)=(u-\ql_{\level{u}})/(\ql_{\level{u}+1}-\ql_{\level{u}})$ be the 
relative distance of $u$ to level $\level{u}+1$. We define the random function
$q_\bql(u):[0,1]\ra\{\ql_0,\ldots,\ql_{s+1}\}$ such that $q_\bql(u)=\ql_{\level{u}}$ with probability $1-\qcoeff(u)$ and 
$q_\bql(u)=\ql_{\level{u}+1}$ with probability $\qcoeff(u)$. Let $q\in\integers_+$ and  $\vbf\in\reals^d$. We define the random quantization of $\vbf$ as follows: 
\begin{align}\nn
 Q_\bql(\vbf):= \|\vbf\|_q\cdot \sbf\odot [q_\bql(u_1),\ldots,q_\bql(u_d)]^\top   
\end{align} where $\odot$ denotes the element-wise (Hadamard) product. 
\end{definition}
We note that $\qbf_\bql=\{q_\bql(u_i)\}_{i=1,\ldots,d}$ are {\it independent} random variables. The encoding  $\ENCODE(\|\vbf\|_q,\sbf,\qbf_\bql):\reals_+\times \{\pm 1\}^d\times \{\ql_0,\ldots,\ql_{s+1}\}^d\ra \{0,1\}^*$ uses a standard floating point encoding with $C_b$ bits to represent the positive scalar $\|\vbf\|_q$, encodes the sign of each coordinate with one bit, and finally applies an $integer$ encoding scheme $\Psi:\{\ql_0,\ql_1,\ldots,\ql_{s+1}\} \to \{0,1\}^*$ to {\it efficiently} encode each quantized normalized coordinate $q_\bql(u_i)$ with the {\it minimum} expected code-length. Depending on how much knowledge of the distribution of the discrete alphabet of levels is known, a particular lossless prefix code can be used to encode $\qbf_\bql$.
In particular, if the distribution of the frequency of the discrete alphabet $\{\ql_0,\ql_1,\ldots,\ql_{s+1}\}$ is unknown but it is known that smaller values are  more frequent than larger values,  Elias recursive coding (ERC) can be used~\citep{Elias}. ERC is a universal lossless integer coding
scheme with a recursive and efficient encoding and decoding schemes, which assigns shorter codes to smaller values. If the distribution of the frequency of the discrete alphabet $\{\ql_0,\ql_1,\ldots,\ql_{s+1}\}$ is known or can be estimated efficiently, we use Huffman coding, which has an efficient encoding/decoding scheme and achieves the {\it minimum expected code-length} among methods encoding symbols separately~\citep{InfTheory}.

The decoding $\DECODE:\{0,1\}^*\to \reals^d$
in \cref{QGEGalg} first reads $C_b$ bits to reconstruct $\|\vbf\|_q$. 
Then it applies $\Psi^{-1}:\{0,1\}^*\to \{\ql_0,\ql_1,\ldots,\ql_{s+1}\}$ to decode the index of the first coordinate, depending on whether the decoded entry is zero or nonzero, it may read one bit indicating the sign, and then proceeds to decode its value. It then decodes the next symbol. The decoding continues mimicking the encoding scheme and finishes when all  quantized coordinates are decoded. Note that the decoding will fully recover $Q_\bql(\vbf)$ because the coding scheme is lossless. One may slightly improve the coding efficiency in terms of the expected code-length by encoding blocks of symbols at the cost of increasing encoding/decoding complexity. We focus on  lossless prefix coding schemes, which encode symbols separately due to their encoding/decoding simplicity~\citep{InfTheory}.

To implement an efficient Huffman code, we need to estimate probabilities w.r.t. the symbols in our discrete alphabet  $\{\ql_0,\ql_1,\ldots,\ql_{s+1}\}$. This discrete  distribution can be estimated by properly estimating the marginal  probability density function (PDF) of normalized coordinates along the lines of \eg~\citep[Proposition 6]{ALQ}.

Given quantization levels $\bql_t$ and the marginal  PDF of normalized coordinates, $K$ processors can construct the Huffman tree in parallel.  A Huffman tree of a source with $s+2$ symbols can be constructed  in time  $O(s)$ through sorting the symbols by the associated probabilities. It is well-known that Huffman codes minimize the expected code-length: 

\bth[{\citealt[Theorems 5.4.1 and 5.8.1]{InfTheory}}]\label{Huffman}
Let $Z$ denote a random source with a discrete alphabet $\Zc$. The expected code-length of an optimal prefix code to compress $Z$ is bounded by 
$H(Z)\leq \E[L]\leq H(Z)+1$ where $H(Z)\leq \log_2(|\Zc|)$ is the entropy of $Z$ in bits. 
\eth

\end{document}